\documentclass[lettersize,journal]{IEEEtran}
\usepackage{amsmath,amsfonts}
\usepackage{algorithmic}
\usepackage{algorithm}
\usepackage{array}
\usepackage[caption=false,font=normalsize,labelfont=sf,textfont=sf]{subfig}
\usepackage{textcomp}
\usepackage{stfloats}
\usepackage{url}
\usepackage{verbatim}
\usepackage{graphicx}
\usepackage{cite}
\usepackage{multirow} % 导入multirow包
\usepackage{caption}
\usepackage{graphicx}
\usepackage{array}
\usepackage{import}
\usepackage{colortbl}
\usepackage{booktabs}
\usepackage{orcidlink} 
\usepackage{booktabs}
\usepackage{multirow}
\usepackage{makecell}
\usepackage{graphicx}

\def\BibTeX{{\rm B\kern-.05em{\sc i\kern-.025em b}\kern-.08em
    T\kern-.1667em\lower.7ex\hbox{E}\kern-.125emX}}
\usepackage{balance}
\begin{document}
\title{FairForensics: Seeing Expressions and Parsing Demographics via Vision-Language Modeling for Generalizable Fair Deepfake Detection}
\author{Yaning Zhang \orcidlink{0000-0001-8442-2777},  Jiao Wu \orcidlink{0009-0004-9787-2389}, Zan Gao \orcidlink{0000-0003-2182-5741}, \emph{Senior Member, IEEE}, Linlin Shen \orcidlink{0000-0003-1420-0815}, \emph{Senior Member, IEEE}, 

\thanks{Corresponding author: Linlin Shen}
\thanks{Yaning Zhang and Jiao Wu are with the Computer Vision Institute,	School of Artificial Intelligence, Shenzhen
	University, Shenzhen, 518060, China. E-mail: zhangyaning0321@163.com; 2400101073@mails.szu.edu.cn}

\thanks{Zan Gao is with the Shandong Artificial Intelligence Institute, Qilu University of Technology (Shandong Academy of Sciences), Jinan, 250014, China, and also with the Key Laboratory of Computer Vision and System, Ministry of Education, Tianjin University of Technology, Tianjin, 300384, China. E-mail: zangaonsh4522@gmail.com }

\thanks{Linlin Shen is with Computer Vision Institute, School of Artificial Intelligence, Shenzhen University, Shenzhen, 518060, China,
	also with National Engineering Laboratory for Big Data System Computing
	Technology, Shenzhen University, and Guangdong Key Laboratory of Intelligent Information Processing, Shenzhen
	University. E-mail: llshen@szu.edu.cn}
}
\markboth{Journal of \LaTeX\ Class Files,~Vol.~18, No.~9, September~2020}%
{How to Use the IEEEtran \LaTeX \ Templates}
\maketitle
\vspace*{-4em}
\begin{abstract}
The challenge of fair deepfake detection (FDD) has attracted increasing attention. Existing fairness-enhanced detectors often suffer from suboptimal generalization to unseen manipulations and fairness across demographic groups, due to the limited exploration in visual modalities alone. They are typically developed and evaluated on demographically imbalanced distributions, resulting in biased predictions toward minority groups. In this paper, we construct a novel demographically balanced FDD benchmark to train and evaluate the fairness of detectors under both balanced and imbalanced population scenarios. Additionally, we introduce a novel expression and demographic perceptual vision-language model, termed FairForensics, for generalizable fair deepfake detection. FairForensics conducts face forgery generalization enhancement and demographic-aware fairness regularization using demographically balanced settings, high-level expression forgery priors, and demographic-aware language prompts. During face forgery generalization enhancement, built upon the novel observation of significant distribution differences between pristine and forged expression vectors, we design an expression encoder to capture high-level expression-guided forgery patterns, and an expression-perceptual visual encoder that integrates global appearance and expression forgery features via an expression injector while mitigating identity-related bias using an identity-aware patch perturbation module. Under demographic-aware fairness regularization, we propose a demographic-guided language encoder to extract population-aware global language embeddings, which boosts the decoupling of forgery-related features from demographic information via vision-language alignment. In addition, we devise a population-aware prototype fairness objective to enforce both inter-class separability and intra-class alignment across demographic subgroups. Extensive experiments on our balanced demographic benchmark and cross-dataset evaluations show that our method achieves the state-of-the-art in terms of generalization and fairness. Our code will be available at GitHub.
\end{abstract}


\vspace*{-1em}
\section{Introduction}
\begin{figure}[t]
	\centering
	\includegraphics[width=\linewidth]{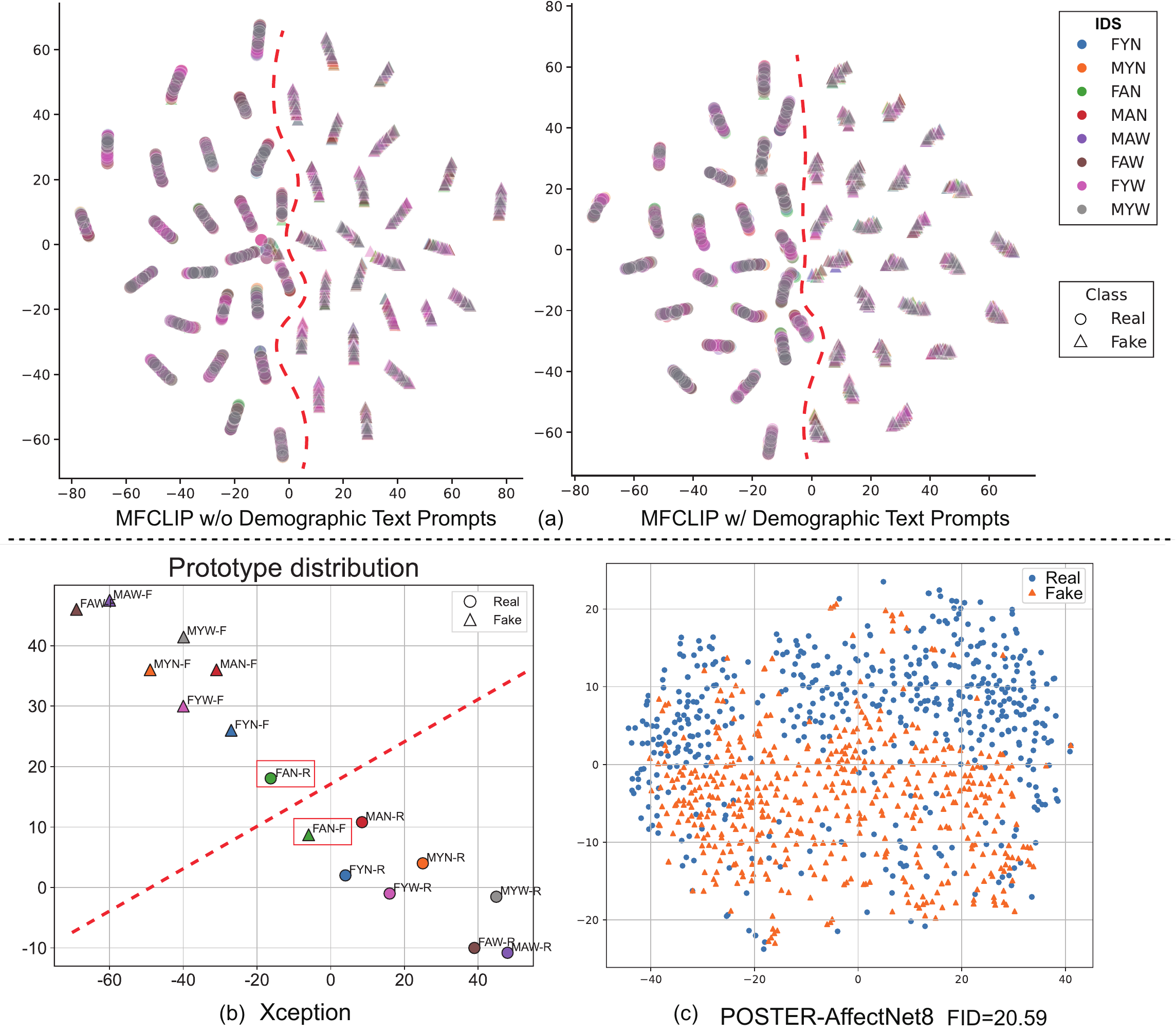}
	\caption{(a) t-SNE visualization of various intersectional demographic subgroups (IDS) embeddings generated by MFCLIP \cite{MFCLIP} without and with demographic text guidance, where IDS is defined by the intersection of multiple demographic attributes, e.g., gender, age, and race. For each IDS, 400 real and 400 fake samples are randomly selected. (b) Prototype distribution generated by Xception, where each point means the class-specific prototype of an IDS. (c) Visualization of facial expression embedding distributions extracted by the pretrained expression estimator \cite{poster} from 500 real and 500 fake faces.
		}\vspace*{-2em}
	\label{fig1}
\end{figure}

\begin{figure}[t]
	\centering
	\includegraphics[width=\linewidth]{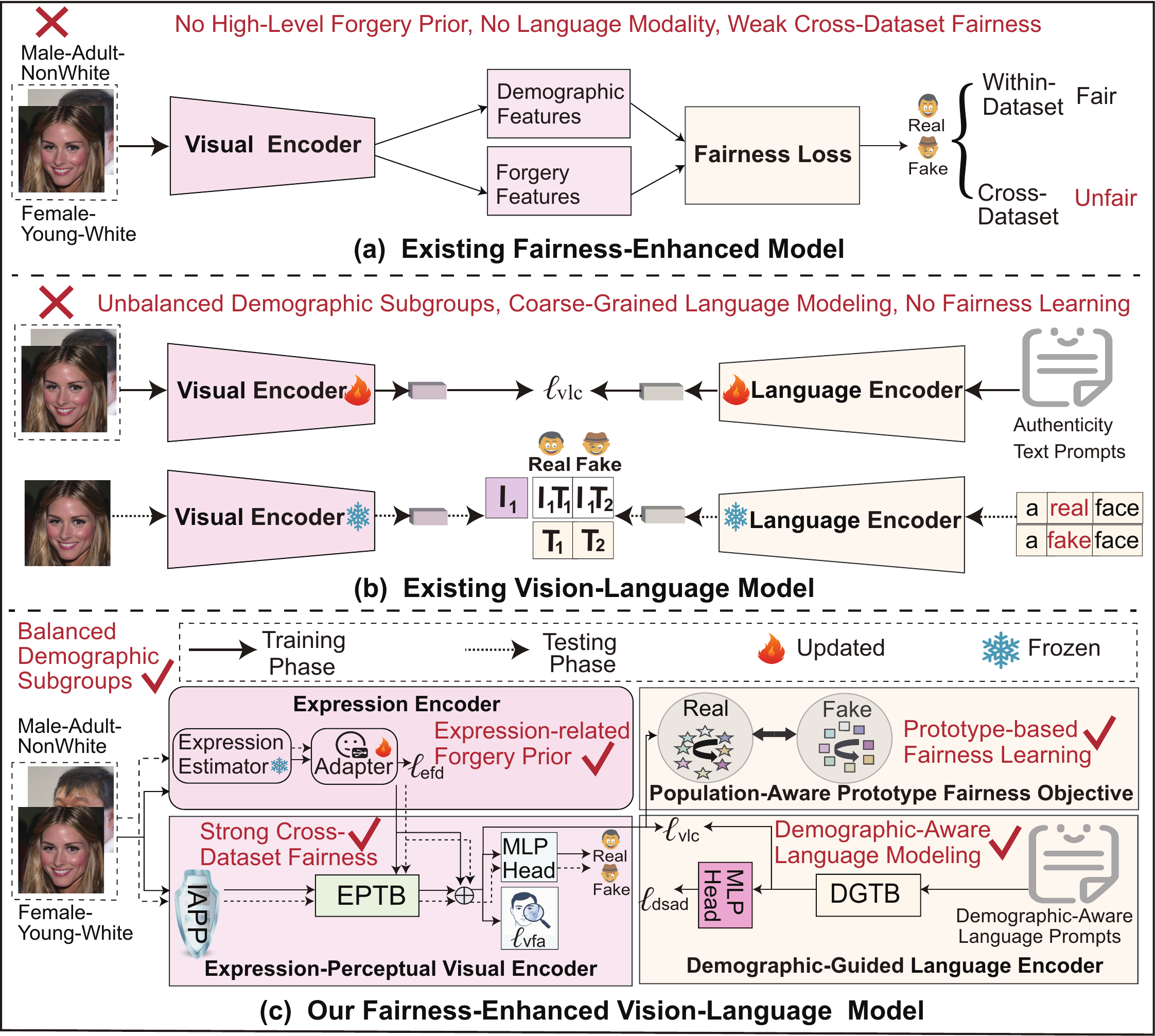}
	\vspace*{-1.5em}
	\caption{(a) Existing fairness-enhanced models mainly rely on demographic features and fairness losses, with insufficient high-level forgery priors, limited exploitation of language modality, and weak cross-dataset fairness. (b) Existing vision-language models introduce authenticity text prompts and coarse-grained language modeling, but fail to explicitly consider demographic-aware language guidance or cross-dataset fairness. (c) Our model incorporates expression-related forgery  priors (EFP), demographic-guided language modeling, and population-aware prototype fairness objective to jointly enhance detection generalization and fairness across IDS.} \vspace*{-2em}
\label{fig2}
\end{figure}

With the rapid advancement of artificial intelligence-generated content, deepfake have emerged as a representative form of face forgery, which aims to synthesize, swap, or manipulate facial images with high visual realism via powerful generative models, such as generative adversarial networks (GANs) \cite{gan} and diffusion models \cite{diff}. Such realistic forgeries increasingly reduce the gap between authentic and manipulated faces, raising security and ethical concerns, including misinformation, identity fraud, and privacy leakage. Therefore, deepfake detection is crucial for identifying forged facial content and mitigating the potential risks of malicious misuse. Despite promising advances, existing deepfake detectors \cite{xception,vgg,CViT} may produce biased predictions across demographic subgroups defined by the intersection of gender, age, and race. These disparities lead to unequal false positive or false negative rates among different populations, exacerbating social bias and compromising the trustworthiness of detectors. For example, in Fig.~\ref{fig1} (b), Xception \cite{xception} tends to confuse real and forged samples from the female-adult-nonwhite (FAN) subgroup, as FAN constitutes a minority group in the deepfake dataset, whereas male-young-white (MYW) samples dominate, biasing the model toward demographic subgroup-specific appearance patterns instead of universal forgery cues. This highlights the necessity of fair deepfake detection (FDD), which intends to achieve accurate and unbiased predictions across diverse demographic groups. Current FDD methods can be broadly categorized into two perspectives: data-centric approaches \cite{gbdf,lin2025ai} and algorithm-centric approaches \cite{crossdf,PFGDD,ding}. The former aims to enhance fairness by creating balanced demographic subgroup distributions from the dataset side. The latter intends to alleviate population-aware bias by disentangling demographic features from domain-agnostic forgery cues at the algorithm level. However, there are some limitations for existing FDD frameworks in terms of generalization and fairness. \textbf{First}, they are typically developed and evaluated on demographically imbalanced distributions, leading to biased predictions and compromised fairness across demographic subgroups. For example, some studies \cite{PFGDD,RSEFFDD,ding} tend to train detectors under the demographically imbalanced scenario and test their generalization and fairness across demographic groups. \textbf{Second}, they are inclined to focus on feature disentanglement within the visual modality and struggle to consider language modeling and distinct high-level forensics prior, which leads to suboptimal generalization and fairness performance. Some works \cite{PFGDD,RSEFFDD,ding} mitigate demographic bias by decoupling population-related representations in the vision domain and regularizing subgroup feature distributions for FDD. \textbf{Finally}, as Fig. \ref{fig2} (b) shows, existing vision-language based models \cite{MFCLIP,clip}, such as CLIP, try to introduce the language modality, but struggle to explicitly conduct fairness learning for generalizable FDD. 

Based on the aforementioned discussion, we aim to study fine-grained demographic-aware language embeddings to enhance the learning of general high-level visual forgery features, to realize generalizable FDD. In this paper, we construct a novel demographically balanced FDD benchmark to train and evaluate the fairness of detectors under both balanced and imbalanced population settings, and propose a fairness-aware vision-language model, namely FairForensics. Unlike prior FDD work, which merely decouples demographic representations from forgery features, our FairForensics method conducts face forgery generalization enhancement and demographic-aware fairness regularization via amplifying high-level expression-aware forgery features and suppressing the forgery-irrelevant identity and demographic cues, for generalizable FDD. In detail, our FairForensics model mainly differs from existing FDD or vision-language based methods in the following aspects (see Fig.~\ref{fig2}): \textbf{First}, during face forgery generalization enhancement, we revisit the role of expression information and make a key observation: There are evident distribution differences (i.e., FID scores) between facial expression embeddings extracted from real and fake images (see Fig.~\ref{fig1} (c)), which introduces previously underexplored yet highly informative signal for distinguishing real and forged images. Motivated by this insight, we reformulate the expression not as an auxiliary cue but as a structured, transferable supervisory signal for generalizable FDD. Thus, we propose an expression encoder to capture comprehensive and high-level expression forgery patterns. \textbf{Second}, we formulate an expression-perceptual visual encoder (EPVE) to explore expression-guided global general appearance forgery traces. However, we note that global features tend to carry manipulation-irrelevant identity semantic information \cite{iil}. To solve this problem, we design a plug-and-play identity-aware patch perturbation module (IAPP) to efficiently suppress forgery-unrelated identity information via introducing identity-aware perturbations at the image patch level. \textbf{Third}, during demographic-aware fairness regularization, we are inspired by the observation that the demographic-aware language guidance mitigates demographic-specific domain shifts (see Fig.~\ref{fig1} (a)). In detail, without population-aware language prompts, demographic subgroup samples tend to gather in respective domains. When the language guidance is involved, various face images from demographic subgroups begin to converge and eventually form a unified cluster. Therefore, we design a demographic-guided language encoder (DGLE) to extract global population-perceptual language features, which facilitates the disentanglement of visual forgery cues from demographic representations via vision-language matching. \textbf{Finally}, since facial feature distributions vary across demographic subgroups even within the same authenticity (real/fake) class (see Fig.~\ref{fig1} (b)), the detector may exploit demographic shortcuts and produce unstable decision boundaries across subgroups. To address this, we introduce a population-aware prototype fairness objective (PPF) to enlarge the separation between real and fake faces and suppress demographic-specific feature shifts within the authenticity class at the prototype level, which boosts our model to learn demographically invariant yet discriminative face forgery features, thus improving both generalization and fairness across diverse subgroups. In summary, the contributions of our work are as follows:

$\bullet$ To the best of our knowledge, we are the first vision-language based work for generalizable FDD under balanced demographic settings, which experimentally shows that balanced population datasets improve cross-domain generalization and demographic-aware text prompts enhance fairness.

$\bullet$ Based on novel findings of high-level expression-related feature differences, we propose a FairForensics model, which enhances face forgery generalization and demographic-aware fairness by amplifying expression-aware forgery features while suppressing forgery-irrelevant identity and demographic cues.  

$\bullet$ We design an innovative plug-and-play identity-aware patch perturbation module to flexibly suppress forgery-unrelated identity cues at the patch level, which could be integrated into transformer-based models to mitigate identity bias with a slight rise in parameters and computational costs.

$\bullet$ Extensive experiments conducted on our benchmark show that our FairForensics method outperforms the state of the art under generalizable FDD scenarios, and models trained on balanced demographic data tend to exhibit diminished fairness disparities than those trained using imbalanced data.

\vspace*{-1em}
\section{Related Work}
\label{sec:formatting}
%-------------------------------------------------------------------------
\subsection{Deepfake Detection and Fairness}
Early CNN-based detectors, such as Xception-based models \cite{DFFD}, mainly capture local texture artifacts and achieve promising performance on seen manipulation types. However, they often lack global reasoning capability, limiting their effectiveness against unseen forgery attacks. To model long-range dependencies, transformer-based detectors \cite{CViT,genface} have been introduced. For example, CViT \cite{CViT} combines convolutional neural networks with vision transformers (ViT) \cite{ViT} to jointly exploit local artifacts and global manipulation traces. Beyond RGB appearance cues, prior-aware methods further explore complementary forensic signals. CAEL \cite{genface} mines multi-grained appearance and edge-aware forgery representations to capture complementary forgery traces. Besides generalization, fairness has become an important issue in face forgery detection. Existing detectors \cite{xception,vgg,RSEFFDD} may exhibit inconsistent performance across demographic subgroups because facial attributes such as sex, age, race, and identity can act as shortcut cues. Some fairness-enhanced methods \cite{PFGDD,RSEFFDD,fairadapter} attempt to reduce demographic bias. Ding et al. \cite{ding} jointly decouple demographic-sensitive channels at the architectural level and align subgroup feature distributions with the global ones. PF-GDD \cite{PFGDD} disentangles demographic and domain-agnostic forgery features and optimizes them within a flattened loss landscape, to achieve cross-domain fairness. Unlike prior methods that rely on visual debiasing or subgroup-level constraints, our model integrates balanced demographic groups, high-level EF priors, demographic-aware language modeling, and prototype-based fairness regularization to achieve generalizable FDD.
%------------------------------------------------------------------------\
\vspace*{-1em}
\subsection{Vision-Language Models}
Vision-language models such as CLIP learn transferable visual-textual representations by aligning image and language embeddings from large-scale image-text pairs, showing strong zero-shot and cross-domain generalization ability. Recent studies have adapted CLIP to deepfake detection by introducing task-specific visual or textual guidance. ForensicAdapter (FA) \cite{forensics} captures distinctive blending boundary artifacts in forged faces via an adapter, and enhance CLIP visual tokens via the interaction to enhance knowledge transfer between CLIP and the adapter. MFCLIP \cite{MFCLIP} integrates fine-grained noise forgery features with global image forgery prompts and improves cross-generator generalization through adaptive vision-language alignment. FairAdapter \cite{fairadapter} adapts CLIP with fairness-aware semantic enhancement and category-balanced optimization to address content-level fairness in AI-generated image detection. They mainly focus on improving generalization and rarely alleviate demographic bias under cross-domain scenarios. In contrast, our method extends vision-language modeling toward generalizable FDD by introducing high-level EFP, identity-aware patch perturbations, demographic-guided language prompts, and population-aware prototype fairness optimization. This enables the model to learn forgery-sensitive yet population-agnostic features, boosting both cross-domain detection generalization and fairness.

\begin{figure*}[t!]%\citep{wodajo2021deepfake,coccomini2022combining}
	\centering
	\includegraphics[width=\linewidth]{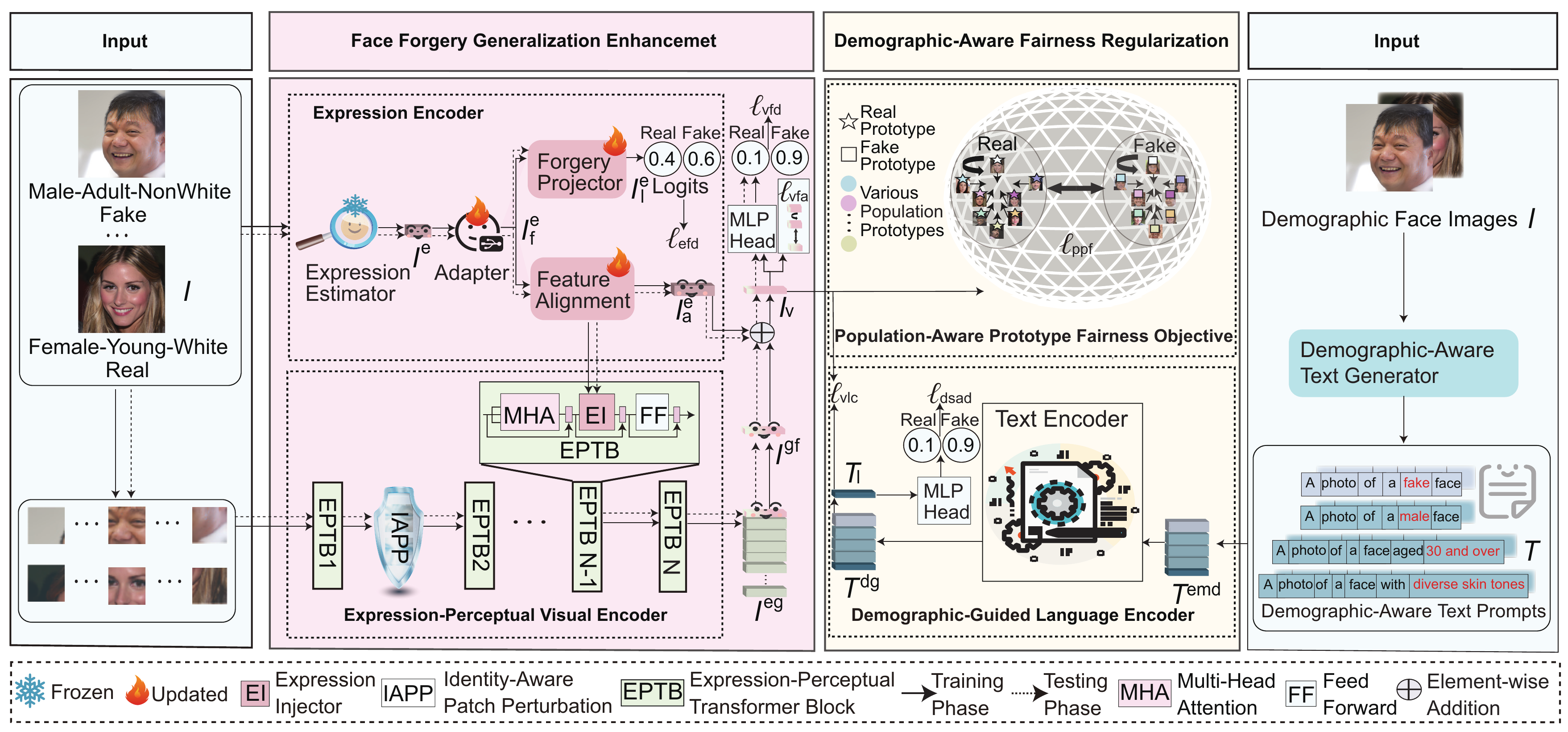} 
	\caption{The workflow of our FairForensics model. After obtaining multiple patches and demographic texts of the input face image, EE is employed to generate EFP features, and DGLE is used to create demographic-aware language embeddings. Then the EFP features and multiple patches are passed into EPVE to suppress forgery-unrelated identity semantic information via IAPP and generate expression-aware general appearance forgery patterns. They are then fused with EFP embeddings to derive visual counterfeit features, which are then fed into the MLP head to make predictions, VFA loss to amplify forgery features, VLC loss to conduct vision-language matching, and the PPF objective to improve fairness. During testing, the trained EE and EPVE module are applied to achieve FDD.  } \vspace*{-1em}
	\label{fig3}
\end{figure*}
\vspace*{-1em}
\section{Methodology}

\subsection{Method Overview}

\noindent{\bfseries Problem Definition.}
For generalizable FDD, given a seen domain $(X_\text{s}, Y_\text{s}, G_\text{s})$ and an unseen domain $(X_\text{u}, Y_\text{u}, G_\text{u})$ with different input distributions, they share the same image authenticity and demographic label spaces, i.e., $X_\text{s} \neq X_\text{u}$, $Y_\text{s}=Y_\text{u}$, and $G_\text{s}=G_\text{u}$. Here, $X$ denotes the input distribution, $Y$ is the image authenticity class label space, and $G$ means the demographic subgroup label space, and we simplify the definition of the image authenticity class as the class. We aim to learn a deepfake detector from the seen domain that can accurately distinguish real and fake faces from the unseen domain while maintaining consistent performance across demographic subgroups. In our work, we employ the training set $(I_\text{s}, y_\text{s}, g_\text{s}) \in \mathcal{D}_\text{tra}$ from the seen domain to train our model, and evaluate it on the testing set $(I_\text{u}, y_\text{u}, g_\text{u}) \in \mathcal{D}_\text{tes}$ from the unseen domain. Here, $I$ denotes the input face image, $y \in {[0,1]^T,[1,0]^T}$ is the real/fake one-hot label, and $g \in {\mathbf{e}_i \mid \mathbf{e}_i=(0,0,\ldots,1,\ldots,0), 1\leq i\leq K}$ is the population subgroup label, where $\mathbf{e}_i \in \mathbb{R}^{K}$ is the $i$-th one-hot vector and $K$ is the number of  intersectional demographic subgroups (IDS). In this paper, each IDS is balanced and labeled with the intersection of population attributes (e.g., sex, age, and race).

{\bfseries\setlength\parindent{0em} Framework.} Fig.~\ref{fig3} illustrates the overall framework of our method from a feature decomposition perspective. Given a face image $I$, its latent representation can be decomposed into forgery-relevant and forgery-irrelevant features. The former captures intrinsic forgery evidence, while the latter mainly includes identity-related characteristics and demographic attributes. Since forgery-irrelevant cues may be exploited as shortcuts, they can degrade generalization of detectors and cause biased predictions across IDS. To address this issue, our framework contains two complementary stages: \emph{\textbf{Face Forgery Generalization Enhancement}} and \emph{\textbf{Demographic-Aware Fairness Regularization}}. In the first stage, we enhance manipulation-relevant representations by extracting high-level EFP cues with an expression encoder (EE) and injecting them into the expression-perceptual visual encoder (EPVE), where an identity-aware patch perturbation (IAPP) module selectively regularizes identity-sensitive patch tokens, to reduce the model's reliance on identity-related shortcuts. In the second stage, we suppress demographic-related bias through language- and prototype-level fairness learning. Specifically, we construct demographic-aware text prompts at the authenticity, sex, age, and race level, and encode them with a demographic-guided language encoder (DGLE) to capture global population-perceptual language features, which facilitates the disentanglement of visual forgery cues from demographic features via vision-language matching. Furthermore, a population-aware prototype fairness (PPF) objective enlarges the separation between real and fake prototypes while aligning IDS prototypes within the same authenticity class. By jointly optimizing these stages, our method learns forgery-discriminative yet identity- and demographic-invariant representations for generalizable FDD.
\vspace*{-1em}
\subsection{Face Forgery Generalization Enhancemet}

{\bfseries\setlength\parindent{0em} Expression encoder.} Unlike existing detectors \cite{Freq,implicit,Lin} that tend to introduce discriminative priors such as landmark, identity, or frequency characteristics, we focus on high-level EFP features since there are evident expression embedding  distribution differences between real and fake images. We aim to utilize the discriminative and general high-level EFP features to extract robust and common forgery patterns, thus facilitating generalizable FDD. 

To capture discriminative and general EFP patterns and adapt expression features to our FDD domain, we design the EE, which consists of a pre-trained expression estimator POSTER \cite{ETHXGaze} using AffectNet8, an adapter, a forgery projector, and a feature alignment module. In Fig.~\ref{fig3}, given a batch of $b$ input facial image $I$, following POSTER, we conduct the data normalization and then utilize the frozen pre-trained POSTER \cite{ETHXGaze} to obtain expression embeddings $I^\text{e}\in\mathbb{R}^{b\times 8}$. To adapt expression information to our FFD domain, we propose the adapter with a fully connected layer to capture EFP features ${I}^\text{e}_\text{f}$. i.e., ${I}^\text{e}_\text{f} ={I}^\text{e}W^{\textrm{e}}$, where $W^{\textrm{e}}$ is the trainable parameter of the adapter. It is then fed into a forgery projector with an MLP head to derive the logits (real and fake) ${I}^\text{e}_\text{l}\in\mathbb{R}^{b\times 2}$, which are supervised by expression-aware forgery detection loss to push the adapter to mine EFP patterns. To perform feature alignment and integration with the visual forgery features derived from the expression-perceptual visual encoder  (EPVE), ${I}^\text{e}_\text{f}$ is then transmitted to the feature alignment module with a fully connected layer to obtain ${I}^\text{e}_\text{a}\in\mathbb{R}^{b\times d}$, where $d$ denotes the feature dimension. $I^\text{e}_\text{a}$ is then fed into the EPVE to explore expression-guided common forgery representations.

{\bfseries\setlength\parindent{0em} Expression-perceptual visual encoder.} 
Since there are obvious texture differences between real and fake images \cite{MAT}, and appearance images provide rich texture details, we propose to integrate global appearance manipulation features with EFP ones, to capture comprehensive expression-perceptual global counterfeit embeddings. However, the global appearance features are inclined to contain the forgery-irrelevant identity semantic information \cite{iil}, we design the identity-aware patch perturbation (IAPP) module to alleviate the interference of forgery-irrelevant but identity-related cues. Unlike existing multi-domain methods that are typically limited to local prior interactions, we devise the expression injector (EI) to capture global and diverse correlations between appearance and EFP embeddings. Unlike vanilla CLIP image encoder that only includes the multi-head attention (MHA) and feed forward (FF) layer in each transformer block, we propose the expression-perceptual visual encoder (EPVE), which adds the proposed IAPP and  EI to mine diverse and global expression-guided general forgery embeddings. As Fig.~\ref{fig3} shows, EPVE consists of $N$ expression-perceptual transformer blocks (EPTB) $\text{TB}_j^\text{i}$, $j=1,2,…, N$, and the first EPTB is followed by a IAPP \cite{lora} to mitigate forgery-irrelevant identity representations.  In detail, given $I$, we split it into $m$ non-overlapping square patches, and they are then flattened and projected to 2D token sequences $I^\text{i}\in\mathbb{R}^{b\times m\times d}$ with the dimension of $d$ along the channel. Thereafter, $I^\text{i}$ is appended with a learnable class token to study global forgery features, and then added with a learnable position embedding $P_\text{i}\in\mathbb{R}^{b\times (m+1)\times d}$ to introduce the position information. That is,
$	I_1^\text{evt}=I^\text{i} \oplus P_\text{i}$, where $\oplus$ is the element-wise addition. After that, $I_1^\text{evt}$ and the expression-aware forgery features $I^\text{e}_\text{a}$ are sequentially transmitted into the first EPTB, the IAPP, and $N-1$ EPTBs, i.e.,
\begin{align}
	\text{EPVE}(I_1^{\text{vet}})
	&=  \text{TB}^\text{i}_{N}\circ   \cdots \text{TB}_2^\text{i}   \circ \text{IAPP}  \circ\text{TB}_1^\text{i}(I_1^\text{evt},I^\text{e}_\text{a} ) \nonumber \\
	&= \text{TB}^\text{i}_{N}\circ   \cdots \text{TB}_2^\text{i}   \circ \text{IAPP} \circ  (I_{2}^\text{evt},I^\text{e}_\text{a})
	\nonumber \\
	&= \cdots =\ \text{TB}^\text{i}_{N}(I_{N}^\text{evt},I^\text{e}_\text{a})=I^\text{eg},
\end{align}
where $\circ$ denotes the function decomposition. To fully dig into diverse and comprehensive expression-aware appearance forgery features, each EPTB $\text{TB}_j^\text{i}$ includes the $\text{MHA}_j^\text{i}$, $\text{GI}_j$, and $\text{FF}_j^\text{ i}$ layer. Specifically, as Fig.~\ref {fig3} illustrates, in the $j$-th EPTB ${\text{TB}_j^\text{i}}$, the expression-perceptual appearance forgery embedding $I_{j}^\text{evt}$ is first transmitted to the $\text{MHA}_j^\text{i}$ to extract global and diverse facial manipulated features $I_j^\text{tok}\in\mathbb{R}^{(m+1)\times d}$. To capture abundant and global expression-aware appearance forged patterns, we design the expression injector (EI). In $\text{EI}_j$, to more efficiently integrate appearance-expression forgery traces, we first decompose $I_j^\text{tok}$ into the class token $I_j^\text{cls}\in\mathbb{R}^{b\times 1\times d} $ and patch tokens $I_j^\text{pat} \in\mathbb{R}^{b\times m\times d}$. To reduce computational overhead, we use only the class token as the query to interact with expression-aware forgery features.
\begin{align}
	Q_j &= I_j^\text{cls}W^\text{que}_j, K_j =I^\text{e}_\text{a}W^\text{key}_j, V_j= I^\text{e}_\text{a}W^\text{val}_j.
\end{align}
To model comprehensive and global correlations between appearance-level and expression-related forgery features, we perform parallel appearance-expression interaction,
\begin{align}
	I_{j}^\text{glo} &=\delta(\frac{Q_{j}K_{j}^T}{\sqrt{\frac{d}{h}}})V_{j},
\end{align}
where $\delta$ is the softmax function, and $h$ is the number of heads. Thereafter, it is fed into a FF layer and then added with $I_j^\text{cls}$ to enhance expression-aware counterfeit traces, i.e., $I_{j}^\text{add} = I_{j}^\text{glo}W^\text{fc}_j+I_j^\text{cls}$, where $W^\text{fc}_j\in\mathbb{R}^{d\times d}$ is the learnable weight of the FF layer. It is then concatenated with the patch token $I_j^\text{pat}$, allowing the forgery knowledge learned from the expression token to be propagated to its corresponding patch tokens, thus boosting the interaction between appearance patches and expression features, i.e., ${I}_j^\text{ae} = [I_{j}^\text{add}||I_j^\text{pat}]\in\mathbb{R}^{ b\times(m+1)\times d}.$ Finally, ${I}_j^\text{ae}$ is transmitted into a $\text{FF}_j^\text{ i}$ layer with a fully connected layer to yield $I_{j+1}^\text{evt}=\text{FF}_j^\text{i}({I}_j^\text{ae})+{I}_j^\text{ae}\in\mathbb{R}^{b\times(m+1)\times d}$. 

%Finally, EPVE yields global and common expression-aware appearance forgery traces $I_\text{g}^\text{ae}\in\mathbb{R}^{1\times d} $ using the class token in $ I^\text{eg}$ generated by the last EPTB. 

\begin{figure}[t!]%\citep{wodajo2021deepfake,coccomini2022combining}
	\centering
	\includegraphics[width=\linewidth]{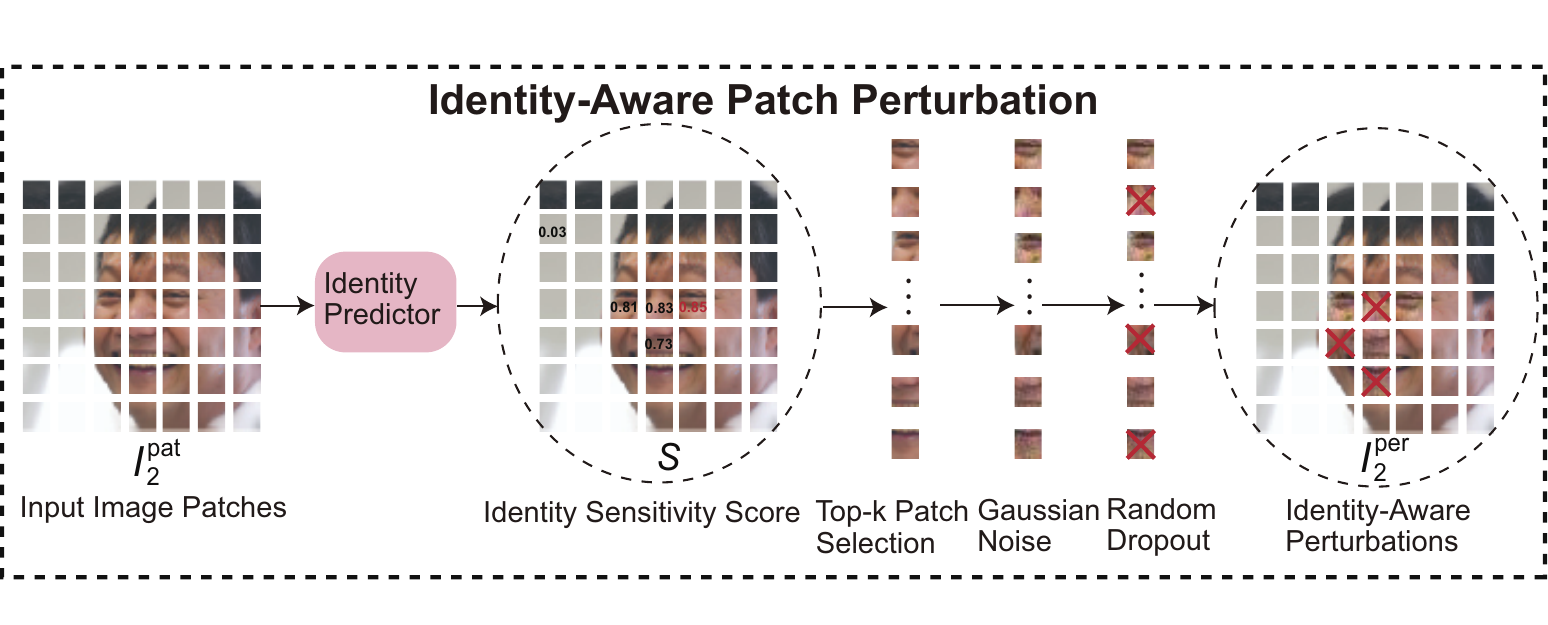}	
	\caption{The pipeline of IAPP.
 }\vspace*{-1em}  \label{fig4}  
\end{figure}

{\bfseries\setlength\parindent{0em}  Identity-aware patch perturbation module.}
Current FDD methods tend to learn discriminative yet shortcut-prone visual forgery cues. In face images, identity-related local regions may dominate the representation learning process, causing the detector to rely on person-specific facial characteristics rather than intrinsic forgery traces. Such identity-biased features can be effective on seen identities or source datasets, but they may degrade the generalization ability when encountering unseen identities, unseen generators, or cross-dataset scenarios. Therefore, it is important to weaken the over-dependence on dominant identity-sensitive patches, and encourage the model to discover robust and forgery-relevant visual evidence from diverse facial regions. To this end, we introduce an IAPP module. Specifically, given the first EPTB output tokens $I_2^\text{evt} \in \mathbb{R}^{b \times (m+1) \times d}$,  we first split it into the class token $I_2^\text{cls}\in\mathbb{R}^{b\times 1\times d} $ and patch tokens $I_2^\text{pat} \in\mathbb{R}^{b\times m\times d}$. Instead of randomly perturbing all image tokens, in Fig.~\ref{fig4}, IAPP first estimates the identity sensitivity of each patch token and selectively regularizes the most identity-responsive patches. In detail, $I_2^\text{cls}$ is preserved to maintain global semantic aggregation, for each patch token in $I_2^\text{pat}$, a lightweight identity scoring head is used to estimate its identity sensitivity:
\begin{equation}
	S = \sigma \left( f_{\mathrm{id}}(I_2^\text{pat}) \right) \in \mathbb{R}^{b \times m},
\end{equation}
where $f_{\mathrm{id}}(\cdot)$ denotes a two-layer MLP and $\sigma(\cdot)$ is the sigmoid function. Based on the obtained identity sensitivity scores, we select the top-$k$ most identity-responsive patches:
$
k = \max(1,\lfloor rm \rfloor),
$
where $r$ is the predefined selection ratio. A binary mask $M \in \mathbb{R}^{b \times m}$ is then generated to indicate the selected patch tokens. During training, IAPP perturbs the selected identity-sensitive patch tokens using two complementary operations. First, Gaussian noise is injected into the selected patch embeddings:
\begin{equation}
	\hat{I_2}^{\mathrm{pat}} = I_2^\text{pat}+ \epsilon \odot M \odot S,
\end{equation}
where $\epsilon \sim \mathcal{N}(0,\sigma^2)$, and the identity sensitivity score $S$ adaptively controls the perturbation strength. Second, patch-level dropout is applied to further suppress part of the selected tokens:
$	I_2^{\mathrm{per}} = \hat{I_2}^{\mathrm{pat}} \odot (1-D),
$
where $D$ denotes the dropout mask generated on the selected identity-sensitive patches. Thereafter, the unchanged class token and the perturbed patch tokens are concatenated to derive
$I^\text{iapp} = [I_2^{\mathrm{cls}} \| I_2^{\mathrm{per}}]$, which is then fed into the next EPTB together with $I^\text{e}_\text{a}$. During inference, both Gaussian noise and patch dropout are disabled, and the original patch representations are directly used. In this way, IAPP serves as a training-time regularization module that reduces identity-biased shortcut learning and improves the generalization ability of the forgery detector.

Finally, EPVT generates the generalizable global forgery features $I^\text{gf} \in \mathbb{R}^{b \times d}$ using the class token in $I^\text{eg}$, where forgery-irrelevant identity features are surpassed, and EFP embeddings are enhanced. $I^\text{gf}$ is then added with expression-perceptual forgery features $I^\text{e}_\text{a}$ to derive the common visual forgery features $I_\text{v}\in \mathbb{R}^{b \times d}$, i.e., $I_\text{v} =  I^\text{gf} \oplus I^\text{e}_\text{a}$.

\vspace*{-1em}
\subsection{Demographic-Aware Fairness Regularization}

\begin{figure}[t!]%\citep{wodajo2021deepfake,coccomini2022combining}
	\centering
	\includegraphics[width=\linewidth]{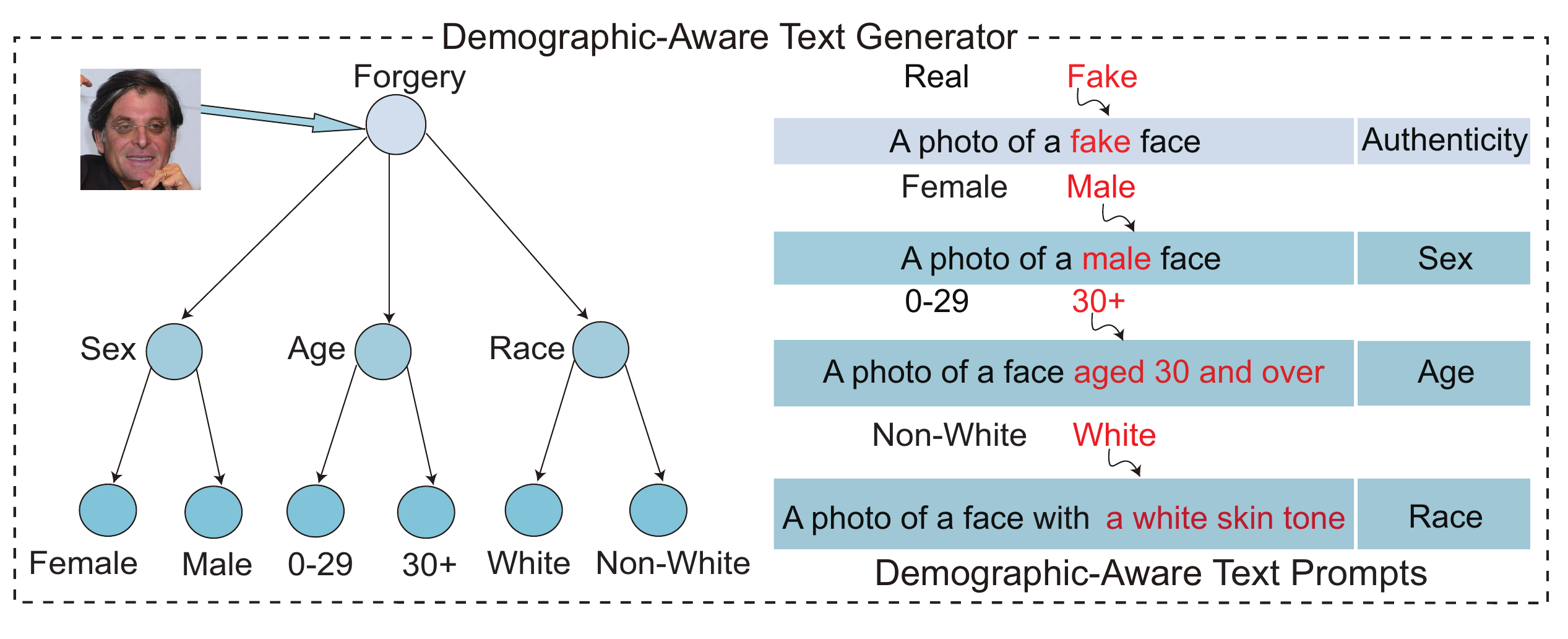} 
	\caption{The architecture of the demographic text generator. }\label{datg}
	\vspace*{-2em}  
\end{figure}
{\bfseries\setlength\parindent{0em}Demographic-aware text generator.} To construct demographic-aware text prompts, we design the demographic-aware text generator (DATG). As shown in Fig.~\ref{datg}, DATG generates corresponding text prompts for each image according to its authenticity and demographic attributes label, thereby building fine-grained demographic-aware image-text pairs. Specifically, we formulate text descriptions from multiple semantic levels. At the authenticity level, DATG describes the authenticity of an image. Regarding the sex level, DATG incorporates the sex attribute and generates prompts such as `a photo of a male face' or `a photo of a female face'. With respect to age level, the prompts are formulated as `a photo of a face aged 0-29' or `a photo of a face aged 30 and over'. DATG further describes the race-related appearance attribute using prompts such as `a photo of a face with a white skin tone' or `a photo of a face with diverse skin tones'. In this way, each facial image is associated with a set of demographic-aware text prompts from authenticity, sex, age, and race perspectives. They are then fed into the demographic-guided language encoder (DGLE) to learn forgery-discriminative and demographic-aware semantic representations. The joint authenticity and demographic prompts facilitate bidirectional semantic disentanglement. Demographic prompts help isolate manipulation-irrelevant population cues from forgery representations, while authenticity ones guide visual features toward manipulation-relevant real/fake semantics. In this way, the model suppresses demographic shortcut learning and strengthens forgery-discriminative supervision, achieving a better trade-off between detection accuracy and fairness.

{\bfseries\setlength\parindent{0em}Demographic-guided language encoder.} To extract hierarchical demographic-aware language embeddings for fairness enhancement, we devise the DGLE, which includes a text encoder (TE) with $L$ transformer blocks ${\text{TB}_j^\text{t}}$, $j=1,2,…, L$. In Fig.~\ref{fig3}, given a batch of demographic-aware text prompts $T$, each of which $\{T_{n}\}_{n=1}^4$ contains four sentences across authenticity, sex, age, and race levels $\{T_{ni}^\text{sen}\}_{i=1}^4$. For each sentence $T_{ni}^\text{sen}$, we utilize the tokenizer \cite{clip} to generate a sequence of word tokens $T_{ni}^\text{tok}\in\mathbb{R}^{77}$, which is then projected to the word embedding vector $T_{ni}^\text{emb}\in\mathbb{R}^{77\times s} $ to derive ${\ T}_i^\text{emb}\in\mathbb{R}^{b\times77\times s}$, where $s$ is the feature dimension. We concatenate four sentences to derive ${ T}^\text{emb}\in\mathbb{R}^{b\times308\times s} $, which is then added with position embedding $P_\text{t}\in\mathbb{R}^{b\times308\times s}$, i.e., $T_1^\text{tra}={\ T}^\text{emb}+P_\text{t}$. Thereafter, it is sequentially fed into $L$ blocks to yield demographic-aware global language features $T_\text{l}\in\mathbb{R}^{b\times s} $, which are then fed to the vision-language contrastive optimization loss (see Section~\ref{lo}) to enhance the fairness of our model. 

\vspace*{-1em}

\subsection{Loss Function}\label{lo}
%\clearpage
%\setcounter{page}{1}
%\maketitlesupplementary

{\bfseries\setlength\parindent{0em} Expression-aware forgery detection loss.}
To mine expression-aware forgery patterns for FDD, $I^\text{e}_\text{a} \in\mathbb{R}^{b\times d}$ is fed into the MLP head to derive the predicted logits $e_\text{pre} \in\mathbb{R}^{b\times 2}$, we devise the EFD loss as follows:
\begin{align}
	\mathcal{L}_\text{efd}=\ \frac{1}{b}\sum_{u=1}^{b}{-{y^u}^T\text{log}(}e_\text{pre}^u),
\end{align}
where $b$ denotes the number of samples in a batch, and $u \in b$ is the index of samples.

{\bfseries\setlength\parindent{0em} Visual forgery detection loss.}
To extract discriminative visual forgery features for FDD, $I_\text{v}\in\mathbb{R}^{b\times d}$ is fed into the MLP head to produce the predicted logits $y_\text{pre} \in\mathbb{R}^{b\times 2}$. Like EFD, we devise the VFD loss. 

{\bfseries\setlength\parindent{0em}Visual forgery amplification loss.}
Unlike traditional methods, which tend to utilize image reconstruction for visual forgery feature enhancement, inspired by CLIP, we devise the VFA loss to conduct the vision contrastive learning for visual forgery amplification.  Mathematically, for the vision-vision feature pairs $\{(I_\text{v}^{u}, I_\text{v}^{u})\}_{u=1}^{b}$, we calculate the vision-to-vision cosine similarity vector as follows:

\begin{align}
	E_\text{v2v}^{uv}(I_\text{v}, I_\text{v})&=\frac{\text{exp}(\text{C}{(I}_\text{v}^u{,I}_\text{v}^v)/\tau)}{\sum_{v=1}^{b}{\text{exp}(\text{C}{(I}_\text{v}^u{,I}_\text{v}^v)/\tau)}}, 
\end{align}
where $\tau$ is a learnable temperature weight to smooth features, C (·) executes a dot product operation to compute similarity scores, and $u, v=1,2,...,b$ are the indices of the sample. The one-hot label $y_\text{pa}$ of the $u$-th pair is denoted as $y_\text{pa}^u=\{{y_\text{pa}^{uv}}\}_{v=1}^b$, $y_\text{pa}^{uu}=1$,
$y_\text{pa}^{uv,u\neq v}=0$. To pull positive pairs together while pushing negative ones away, the VFA loss is denoted as follows:

\begin{align}
	\mathcal{L}_\text{vfa}
	&=\frac{1}{b}\sum_{u=1}^{b}{-({y_\text{pa}^u})^T\text{log}(}E_\text{v2v}^u).
\end{align}

{\bfseries\setlength\parindent{0em}Population-aware prototype fairness objective.}
For FDD, face samples from different demographic groups may exhibit different feature distributions, even when they belong to the same authenticity class, which may make the detector rely on demographic-specific shortcuts, leading to unstable decision boundaries across IDS. Meanwhile, real and fake samples should remain clearly separable regardless of demographic attributes. To achieve this, we introduce a population-aware prototype fairness objective (PPF), which regularizes the prototype feature space from both inter-class and intra-class perspectives (see Fig.~\ref{fig3}). 

Specifically, given a mini-batch of visual features $I_\text{v}=\{I_\text{v}^u\}_{u=1}^{b}$, authenticity class labels $Y=\{y^u\}_{u=1}^{b}$, and demographic group labels $G=\{g^u\}_{u=1}^{b}$, where $y^u \in \{0,1\}$ denotes real or fake, we first normalize each feature as $	\tilde{I_\text{v}^u} = \frac{I_\text{v}^u}{\|I_\text{v}^u\|_2}.
$ Unlike existing fairness constraints, we perform prototype-level regularization rather than sample-level one, which provides a more stable estimation of subgroup feature distributions. For each demographic group $g$ and authenticity class $y$, we compute the class-conditional subgroup prototype by averaging the features belonging to the same group and class:
\begin{equation}
	\mu_g^y =
	\frac{1}{|\mathcal{D}_g^y|}
	\sum_{u \in \mathcal{D}_g^y} 	\tilde{I_\text{v}^u},
\end{equation}
where
\begin{equation}
	\mathcal{D}_g^y = \{u \mid g^u=g, y^u=y\}.
\end{equation}

To ensure that real and fake representations are separated across IDS, we explicitly considers demographic groups and image authenticity classes jointly, enabling the model to learn fairer decision boundaries across subgroups. we define the group-class margin loss as
\begin{equation}
	\mathcal{L}_{\mathrm{mar}}
	=
	\frac{1}{|\mathcal{G}_0||\mathcal{G}_1|}
	\sum_{g \in \mathcal{G}_0}
	\sum_{g' \in \mathcal{G}_1}
	\left[
	\max
	\left(
	0,
	mar - \left\|\mu_g^0 - \mu_{g'}^1\right\|_2
	\right)
	\right]^2,
\end{equation}
where $mar$ is the predefined margin, $\mathcal{G}_0$ denotes the set of IDS containing real samples, and $\mathcal{G}_1$ means the set of IDS containing fake samples. $	\mathcal{L}_{\mathrm{mar}}$ pushes real and fake prototypes to be separated by at least a margin $mar$, even when they come from different IDS.

To reduce demographic-induced feature shifts within the authenticity class, we introduce the class-conditional subgroup alignment loss, which combines inter-class separation and intra-class alignment to preserve the discriminability of detectors between real and fake faces, and suppress demographic-specific feature shifts within the class,
\begin{equation}
	\mathcal{L}_{\mathrm{align}}
	=
	\frac{1}{2}
	\sum_{y \in \{0,1\}}
	\frac{1}{|\mathcal{G}_y|(|\mathcal{G}_y|-1)}
	\sum_{\substack{g,g' \in \mathcal{G}_y \\ g \neq g'}}
	\left\|
	\mu_g^y - \mu_{g'}^y
	\right\|_2^2,
\end{equation}
where $\mathcal{G}_y$ is the set of IDS that contain samples of class $y$. $\mathcal{L}_{\mathrm{align}}$ pulls subgroup prototypes of the same class closer, so that real samples from different groups share similar real representations, and vice versa. The PPF loss is formulated as follows:
\begin{equation}
	\mathcal{L}_{\mathrm{ppf}}
	=
	\mathcal{L}_{\mathrm{mar}}
	+
	\mathcal{L}_{\mathrm{align}}.
\end{equation}

{\bfseries\setlength\parindent{0em}Vision-language contrastive optimization.}
Like vanilla CLIP \cite{clip}, we introduce the vision-language contrastive loss $\mathcal{L}_\text{vlc}$ to conduct the vision-language alignment between $I_\text{v}$ and $T_\text{l}$, which benefits fairness by introducing explicit demographic-aware semantic supervision. Since the textual features contain both authenticity and demographic information, aligning visual embeddings with these language features encourages the model to distinguish real and fake faces under different demographic contexts. For example, for real visual features $I_\textrm{vA}^\textrm{r}$ and fake ones $I_\textrm{vA}^\textrm{f}$ from the same demographic subgroup $\textrm{A}$, their corresponding language embeddings are $T_\textrm{lA}^\textrm{r}$ and $T_\textrm{lA}^\textrm{f}$, the pairs $(I_\textrm{vA}^\textrm{r},T_\textrm{lA}^\textrm{r})$ and $(I_\textrm{vA}^\textrm{f},T_\textrm{lA}^\textrm{f})$ are positives, whereas $(I_\textrm{vA}^\textrm{r},T_\textrm{lA}^\textrm{f})$ and $(I_\textrm{vA}^\textrm{f},T_\textrm{lA}^\textrm{r})$ are negatives. If the model over-relies on demographic cues, the negative pairs may still obtain high similarity because they share the same demographic semantics, which disrupts correct real/fake alignment. Thus, to optimize $\mathcal{L}_\text{vlc}$, the model is encouraged to suppress demographic-related features and amplify manipulation-relevant features, which reduces demographic shortcut learning and promotes more consistent forgery features across IDS, thus improving fairness.

{\bfseries\setlength\parindent{0em} Demographic-aware semantic authenticity discrimination.} Considering that demographic-aware global language features $T_\text{l} $ mainly contain two types of senmatic signals: authenticity and demographic-aware embeddings. To enhance the authenticity language feature for visual forgery patterns enforcement under the guidance of $\mathcal{L}_\text{vlc}$, $T_\text{l}\in\mathbb{R}^{b\times d}$ is fed into the MLP head to produce the predicted logits $T_\text{pre} \in\mathbb{R}^{b\times 2}$. We design the DSAD loss as follows:
\begin{align}
	\mathcal{L}_\text{dsad}=\ \frac{1}{b}\sum_{u=1}^{b}{-{y^u}^T\text{log}(}T_\text{pre}^u).
\end{align}

The overall training objective is defined as:
\begin{equation}
	\mathcal{L}
	=
	\mathcal{L}_{\mathrm{vfd}}
	+
	\mathcal{L}_{\mathrm{efd}}	+
	\mathcal{L}_{\mathrm{vfa}}+\lambda
	\mathcal{L}_{\mathrm{ppf}}+
	\mathcal{L}_{\mathrm{dsad}}+
	\mathcal{L}_{\mathrm{vlc}},
\end{equation}
where $\lambda$ is a trade-off hyperparameter that controls the contribution of $\mathcal{L}_{\mathrm{ppf}}$ to the overall objective.

\begin{figure}[t!]%\citep{wodajo2021deepfake,coccomini2022combining}
	\centering
	\includegraphics[width=\linewidth]{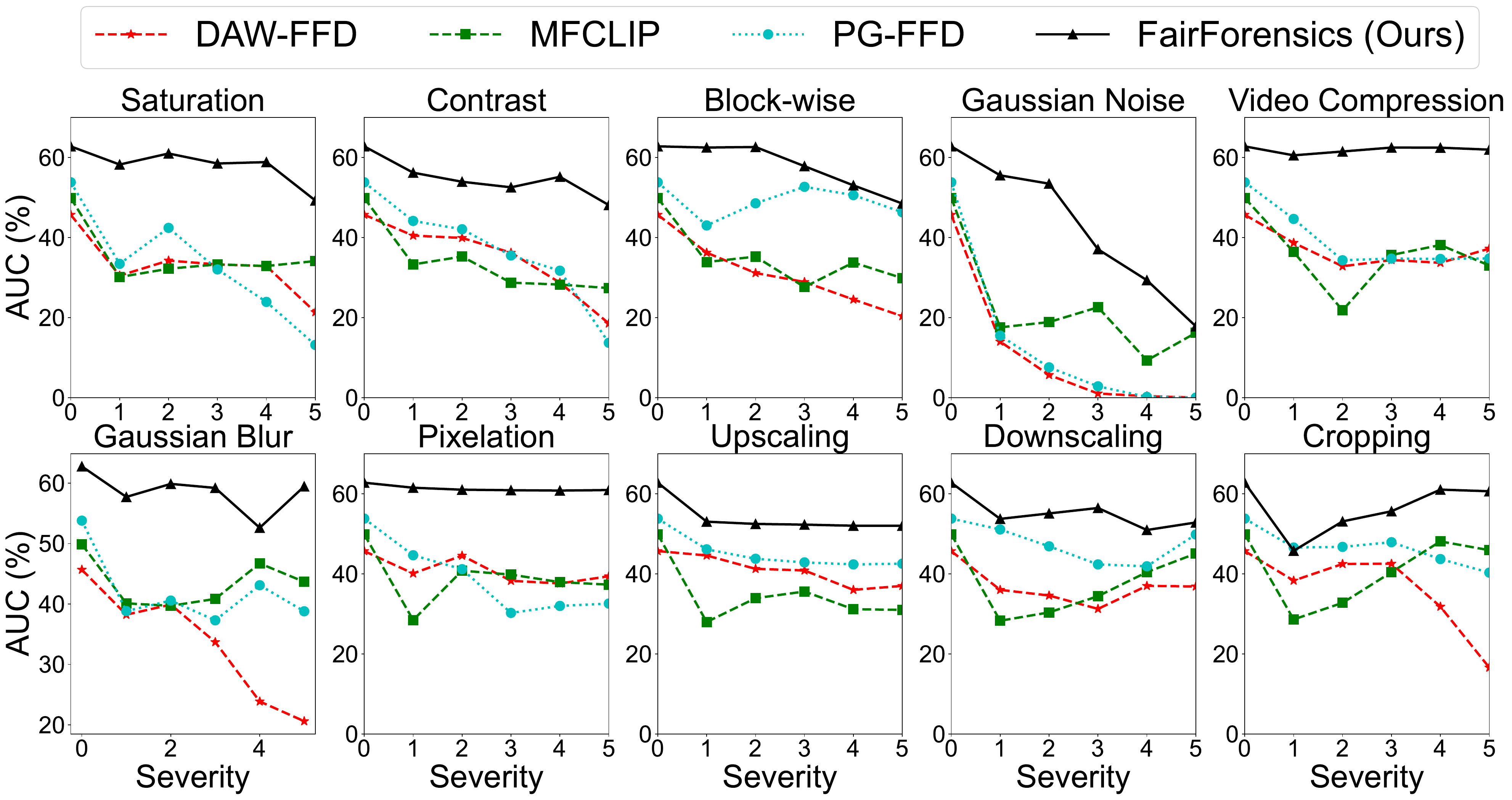} 
	\caption{Resilience to unseen image distortions. }\label{robust2}
	\vspace*{-2em}
\end{figure}

\vspace*{-1em}
\section{Experiments}
\vspace*{-1em}
\subsection{Experiment Setup}

{\bfseries\setlength\parindent{0em} Dataset collection and protocol.}
As shown in Table~\ref{tab1}, we construct a balanced demographic dataset based on GenFace \cite{genface}.  Each image is annotated with an authenticity class label and an intersectional demographic subgroup (IDS) label. The former denotes the binary category, where 0 indicates real and 1 indicates fake. To analyze demographic fairness, we divide the dataset into eight IDS, including male-youth-white (MYW), female-youth-white (FYW), male-adult-white (MAW), female-adult-white (FAW), male-adult-nonwhite (MAN), female-adult-nonwhite (FAN), male-youth-nonwhite (MYN), and female-youth-nonwhite (FYN), according to the intersection of sex, age, and race, where sex includes male and female, age includes 0--29 (Youth) and 30+ (Adult), and race contains White and Non-White. The demographic attributes are automatically annotated using the FairFace algorithm \cite{fairface}. The IDS labels from 0 to 7 correspond to eight demographic subgroups. For each subgroup, we keep the number of samples strictly balanced, with 3,212 images for training, 383 images for validation, and 468 images for testing in both real and fake categories. Our demographically balanced data partitioning follows the original GenFace protocol. This balanced protocol enables a fair training and evaluation across demographic subgroups.

{\bfseries\setlength\parindent{0em} Baseline methods.} We evaluate state-of-the-art methods on our benchmark. We select various deepfake detectors such as CNN-based (ResNet \cite{resnet}, Xception \cite{xception}, VGG \cite{vgg}),  transformer-based (i.e., ViT \cite{ViT}, CViT \cite{CViT}, and CAEL \cite{genface}), vision-language-based (i.e., CLIP \cite{clip}, MFCLIP \cite{MFCLIP}, and ForensicsAdapter (FA) \cite{forensics}), and fairness-enhanced methods (e.g., DAG-FDD \cite{DNADet},  DAW-FDD \cite{omnidfa}, PG-FDD \cite{cdal}, FairAdapter \cite{fairadapter}, RSEF-FDD \cite{RSEFFDD}, and Ding et al. \cite{DEFAKE}).

{\bfseries\setlength\parindent{0em} Evaluation metrics.} We use the area under curve (AUC) as the detection metric. To comprehensively evaluate demographic fairness, we report four fairness metrics, including the false positive rate gap ($F\textsubscript{FPR}$), mean equalized odds gap ($F\textsubscript{MEO}$), demographic parity gap ($F\textsubscript{DP}$), and overall accuracy equality gap ($F\textsubscript{OAE}$). Lower fairness metrics indicate better fairness, while higher AUC denotes stronger detection performance.

{\bfseries\setlength\parindent{0em}Implementation details.}\label{secA}
We implement the model using PyTorch on a Tesla A100 GPU with 40GB of memory under the Linux system. The number of transformer blocks $N$ and $L$ in FairForensics is set to 12. The feature dimensions $d$ and $s$ are configured to 768 and 512. The patch number $m$, head number $h$, the number of IDS $K$, and batch size $b$ are 64, 8, 8, and 24. We set the predefined selection ratio $r$, margin $mar$, loss term $\lambda$, and the standard deviation of Gaussian noise $\sigma^2$ to 0.5, 0.5, 1.0, and 0.05, respectively. The height and width of images are resized to 224. We conduct the normalization by dividing image pixel values by 255, without any data augmentations. Following POSTER \cite{poster}, the input face image of the expression estimator is normalized by mean [0.485, 0.456, 0.406] and standard deviation [0.229, 0.224, 0.225] for each color channel. Our method is trained with the Adam optimizer using a learning rate of 1e-5 and a weight decay of 1e-4. We use the scheduler to drop the learning rate by ten times every 15 epochs. To ensure a fair comparison, we develop SOTA methods and follow default configurations until convergence. All models are trained and evaluated using the same protocols.

%{\bfseries\setlength\parindent{0em}Implementation details.}
%{\bfseries\setlength\parindent{0em} Implementation details.} E=6 a=1024 d=768,s=512, L=12, N=12 chw 512,7,7 r=8, t=308, m=64,
%We implement the model using PyTorch on a Tesla A100 GPU with 40GB of memory under the Linux system. The number of transformer blocks $E$, $L$ and $N$ in GazeCLIP is set to 6, 12, and 12, respectively. The feature dimensions $a$, $d$, and $s$ are configured to 1024, 768, and 512. The patch number $m$, token number $t$, head number $r$, and batch size $b$ are 64, 308, 8, and 32. We set the channel $c$, height $h$, and width $w$ of feature maps to 512, 7, and 7. We set the LoRA rank and alpha to 8 and 32. The height and width of images are resized to 256. We conduct the normalization by dividing image pixel values by 255, without any data augmentations. Following ETH-Xgaze, the input face image of the gaze estimator is normalized by mean [0.485, 0.456, 0.406] and standard deviation [0.229, 0.224, 0.225] for each color channel. Our method is trained with the Adam optimizer \cite{adam} using a learning rate of 1e-4 and a weight decay of 1e-3. We leverage the scheduler to drop the learning rate by ten times every 15 epochs. To ensure a fair comparison, we develop SOTA methods and follow default configurations until convergence. All models are trained and evaluated using the same protocols. We use the cross-entropy as our DFA and DFD loss function. We use the contrastive loss in vanilla CLIP for vision-language matching.%
\begin{table}[t]
	\centering
	\caption{	Outline of the balanced GenFace dataset construction across intersectional demographic subgroups (IDS). \# denotes the number of samples. IA Label indicates the image authenticity label, where 0 and 1 denote real and fake images, respectively. IDS Label denotes the subgroup label defined by the intersection of sex, age, and race from 0 to 7.
	}\label{tab1}
	\label{tab:genface_split}
	\renewcommand{\arraystretch}{1.08}
	\setlength{\tabcolsep}{1.3pt}
	\scriptsize
	\begin{tabular}{cccccccccc}
		\toprule
	\makecell{Dataset} & 
	\makecell{Authen-\\ticity} & 
	Sex & 
	Age & 
	Race & 
	\makecell{Training\\\#} & 
	\makecell{Val\\\#} & 
	\makecell{Testing\\\#} & 
	\makecell{IA\\Label} & 
	\makecell{IDS\\Label} \\
		\midrule
		GenFace & Real & Male  & 0-29 (Youth)& White     & 3,212 & 383 & 468 & 0 & 0 \\
		GenFace & Real & Female & 0-29 (Youth) & White     & 3,212 & 383 & 468 & 0 & 1 \\
		GenFace & Real & Male   & 30+ (Adult) & White     & 3,212 & 383 & 468 & 0 & 2 \\
		GenFace & Real & Female & 30+ (Adult)  & White     & 3,212 & 383 & 468 & 0 & 3 \\
		GenFace & Real & Male   & 30+ (Adult)  & Non-White & 3,212 & 383 & 468 & 0 & 4 \\
		GenFace & Real & Female & 30+ (Adult)  & Non-White & 3,212 & 383 & 468 & 0 & 5 \\
		GenFace & Real & Male   & 0-29 (Youth) & Non-White & 3,212 & 383 & 468 & 0 & 6 \\
		GenFace & Real & Female & 0-29 (Youth)& Non-White & 3,212 & 383 & 468 & 0 & 7 \\
		\midrule
			\midrule
		GenFace & Fake & Male   & 0-29 (Youth)& White     & 3,212 & 383 & 468 & 1 & 0 \\
		GenFace & Fake & Female & 0-29 (Youth)& White     & 3,212 & 383 & 468 & 1 & 1 \\
		GenFace & Fake & Male   & 30+ (Adult) & White     & 3,212 & 383 & 468 & 1 & 2 \\
		GenFace & Fake & Female & 30+ (Adult)  & White     & 3,212 & 383 & 468 & 1 & 3 \\
		GenFace & Fake & Male   & 30+ (Adult)   & Non-White & 3,212 & 383 & 468 & 1 & 4 \\
		GenFace & Fake & Female & 30+ (Adult)   & Non-White & 3,212 & 383 & 468 & 1 & 5 \\
		GenFace & Fake & Male   & 0-29 (Youth)& Non-White & 3,212 & 383 & 468 & 1 & 6 \\
		GenFace & Fake & Female & 0-29 (Youth) & Non-White & 3,212 & 383 & 468 & 1 & 7 \\
		\bottomrule
	\end{tabular}
\end{table}

\vspace*{-1em}
\subsection{Comparison with the State of the Art}

\begin{table}[t]
	\centering
	\caption{Within-dataset evaluation. Detection and fairness performance (\%) on both balanced and unbalanced IDS testing sets after training using either balanced or unbalanced IDS from GenFace. Lower fairness scores indicate better fairness, and higher AUC indicates better detection performance. \textbf{Bold} and \underline{underline} denote the best and the second-best results.
		.}
	\label{within}
	\renewcommand{\arraystretch}{1.00}
	\setlength{\tabcolsep}{0.2pt}
	\scriptsize
	\begin{tabular}{|c|c|c|ccc|c|ccc|c|}
		\toprule
			\toprule
		\multirow{4}{*}{\makecell{Training\\Set}} 
		& \multirow{4}{*}{\makecell{Model\\Type}}
		& \multirow{4}{*}{Detectors} 
		& \multicolumn{8}{c|}{Testing Set} \\
		\cmidrule{4-11}      
		& & 
		& \multicolumn{4}{c|}{Balanced}                 
		& \multicolumn{4}{c|}{Unbalanced} \\
		\cmidrule{4-11}      
		& & 
		& \multicolumn{3}{c|}{Fairness} 
		& Detect.
		& \multicolumn{3}{c|}{Fairness} 
		& Detect. \\
		\cmidrule{4-11}      
		& & 
		& \textit{F\textsubscript{FPR}$\downarrow$} 
		& \textit{F\textsubscript{MEO}$\downarrow$} 
		& \textit{F\textsubscript{DP}$\downarrow$}  
		& AUC$\uparrow$   
		& \textit{F\textsubscript{FPR}$\downarrow$} 
		& \textit{F\textsubscript{MEO}$\downarrow$} 
		& \textit{F\textsubscript{DP}$\downarrow$}  
		& AUC$\uparrow$ \\
		\midrule
		
		\multirow{16}{*}{Balanced} 
		& \multirow{3}{*}{\makecell{CNN\\-based}} 
		& Xception \cite{xception}
		& 0.37 & 1.60 & 0.80 & 99.98 
		& 0.34 & 1.60 & 10.80 & 99.98 \\
		
		& & ResNet \cite{resnet}
		& 0.35 & 0.43 & 0.21 & \underline{99.99} 
		& \underline{0.02} & 0.49 & 12.64 & 100 \\
		
		& & VGG \cite{vgg}
		& \underline{0.16} & \underline{0.11} & \underline{0.05} & 100 
		& 0.36 & 0.21 & 12.51 & 100 \\
		
		\cmidrule{2-11}      
		
		& \multirow{3}{*}{\makecell{Trans\\-based}} 
			& ViT \cite{ViT} 
		& 1.71 & 2.88 & 1.98 & 88.92 
		& 2.19 & 10.35 & \underline{10.42} & 88.97 \\
		
		& & CViT \cite{CViT}
		& \underline{0.16} & \underline{0.11} & \underline{0.05} & 100 
		& \underline{0.02} & 0.16 & 12.46 & 100 \\
		
		& & CAEL \cite{genface}
		& \underline{0.16} & 0.21 & 0.11 & \underline{99.99}
		& 0.23 & 0.50 & 12.52 & \underline{99.99} \\
		
		\cmidrule{2-11}      
		
		& \multirow{3}{*}{\makecell{VLM\\-based}} 
		& CLIP \cite{clip} 
		& 1.82 & 2.35 & 1.50 & 99.84 
		& 2.31 & 2.43 & 13.72 & 99.87 \\
		
		& & MFCLIP \cite{MFCLIP} 
		& 0.91 & 0.75 & 0.37 & 99.95 
		& 0.84 & 0.93 & 12.72 & 99.93 \\
		
		& & FA \cite{forensics}   
		& 0.21 & 0.85 & 0.53 & 100 
		& 0.17 & 0.90 & 12.52 & 100 \\
		
		\cmidrule{2-11}      
		
		& \multirow{4}{*}{\makecell{Fair\\-based}} 
		& DAG-FDD \cite{daw}  
		& 2.03 & 2.35 & 0.96 & 99.70 
		& 1.13 & 3.19 & 13.04 & 99.70 \\
		
		& & DAW-FDD \cite{daw}  
		& 1.56 & 3.21 & 1.60 & 95.07 
		& 1.89 & 3.56 & 10.58 & 95.30 \\
		
		& & PG-FDD \cite{PFGDD}  
		& 0.56 & 0.21 & 0.11 & 98.01 
		& 0.39 & \underline{0.05} & 12.45 & 98.80 \\

		\rowcolor{cyan!12} & & Ours  
		&\textbf{0.10} & \textbf{0.04} & \textbf{0.02} & \textbf{100}
		& \textbf{0.01} & \textbf{0.04} &\textbf{9.35} & \textbf{100} \\
		
		\midrule

		\multirow{16}{*}{Unbalanced} 
		& \multirow{3}{*}{\makecell{CNN\\-based}} 
		& Xception \cite{xception}
		& 0.64 & 3.10 & 1.55 & 99.99 
		& 0.27 & 1.97 & 13.11 & \underline{99.99}\\
		
		& & ResNet \cite{resnet}
		& 0.43 & 2.24 & 1.01 & 99.98 
		& \underline{0.12} & 1.77 & 12.98 & 99.98 \\
		
		& & VGG \cite{vgg}
		& \underline{0.16} & 0.21 & 0.11 & 100 
		& 0.01 & 0.06 & 12.46 & 100 \\
		
		\cmidrule{2-11}      
		
		& \multirow{3}{*}{\makecell{Trans\\-based}} 
		& ViT \cite{ViT}  
		& 1.82 & 3.74 & 2.35 & 89.95 
		& 2.23 & 10.36 & 10.96 & 90.06 \\
		
		& & CViT \cite{CViT}   
		& \underline{0.16} & \underline{0.11} & \textbf{0.05} & 100 
		& 0.16 & 0.16 & 12.39 & 100 \\
		
		& & CAEL \cite{genface}    
		& \underline{0.16} & 0.53 & 0.32 & 100 
		& 0.16 & 0.25 & 12.32 & 100 \\
		
		\cmidrule{2-11}      
		
		& \multirow{3}{*}{\makecell{VLM\\-based}} 
		& CLIP \cite{clip}   
		& 1.39 & 4.81 & 2.56 & 99.82 
		& 1.31 & 3.77 & 14.26 & 99.85 \\
		
		& & MFCLIP \cite{MFCLIP}   
		& 1.15 & 0.44 & 0.20& 99.91 
		& 0.92 & 0.45 & 12.58& 99.90\\
		
		& & FA \cite{forensics}     
		& 0.43 & 0.53 & 0.32 & 100 
		& 0.25 & 0.47 & 12.39 & 100 \\
		
		\cmidrule{2-11}      
		
		& \multirow{4}{*}{\makecell{Fair\\-based}} 
		& DAG-FDD \cite{daw}   
		& 2.54 & 2.36 & 1.24& 99.93 
		& 0.92 & 2.58 & 13.22& 99.95 \\
		
		& & DAW-FDD \cite{daw}   
		& 2.03 & 3.21 & 1.62 & 95.76 
		& 1.60 & 2.92 & \underline{10.71} & 95.93 \\
		
		& & PG-FDD \cite{PFGDD}   
		& 1.04 & 0.23& 0.40 & 98.85 
		& 0.16 & \underline{0.02}& 12.77& 98.88 \\

		\rowcolor{cyan!12} & & Ours  
		& \textbf{0.10} & \textbf{0.08}& \underline{0.06 }& \textbf{100}
		& \textbf{0.01} & \textbf{0.01}& \textbf{10.24} & \textbf{100} \\
		
		\bottomrule
				\bottomrule
	\end{tabular}
\vspace*{-2em}
\end{table}

\begin{table*}[t]
	\centering
	\caption{Performance comparison under cross-dataset evaluation. Detection and fairness performance (\%) on FF++, Celeb-DF, DFDC, and DF-1.0 after training using our balanced IDS from GenFace. Lower fairness scores indicate better demographic fairness, while higher AUC means better detection performance. \textbf{Bold} and \underline{underline} denote the best and the second-best results.}
	\label{cross}
	\renewcommand{\arraystretch}{1.15}
	\setlength{\tabcolsep}{0.2pt}
	\scriptsize
	
	\begin{tabular}{|c|c|cccc|c|cccc|c|cccc|c|cccc|c|cc|}
		\toprule
			\toprule
		\multirow{3}{*}{\makecell{Model\\Type}}
		& \multirow{3}{*}{Detectors}
		& \multicolumn{5}{c|}{FF++}
		& \multicolumn{5}{c|}{Celeb-DF}
		& \multicolumn{5}{c|}{DFDC}
		& \multicolumn{5}{c|}{DF-1.0}
		& \multirow{3}{*}{\makecell{Params\\(M)}}
		& \multirow{3}{*}{\makecell{FLOPs\\(G)}} \\
		\cmidrule{3-22}
		
		& 
		& \multicolumn{4}{c|}{Fairness}
		& Detect. 
		& \multicolumn{4}{c|}{Fairness }
		& Detect.
		& \multicolumn{4}{c|}{Fairness }
		& Detect.
		& \multicolumn{4}{c|}{Fairness}
		& Detect.
		& & \\
		\cmidrule{3-22}
		
		&
		& \textit{F\textsubscript{FPR}$\downarrow$}
		& \textit{F\textsubscript{MEO}$\downarrow$}
		& \textit{F\textsubscript{DP}$\downarrow$}
		& \textit{F\textsubscript{OAE}$\downarrow$}
		& AUC$\uparrow$
		& \textit{F\textsubscript{FPR}$\downarrow$}
		& \textit{F\textsubscript{MEO}$\downarrow$}
		& \textit{F\textsubscript{DP}$\downarrow$}
		& \textit{F\textsubscript{OAE}$\downarrow$}
		& AUC$\uparrow$
		& \textit{F\textsubscript{FPR}$\downarrow$}
		& \textit{F\textsubscript{MEO}$\downarrow$}
		& \textit{F\textsubscript{DP}$\downarrow$}
		& \textit{F\textsubscript{OAE}$\downarrow$}
		& AUC$\uparrow$
		& \textit{F\textsubscript{FPR}$\downarrow$}
		& \textit{F\textsubscript{MEO}$\downarrow$}
		& \textit{F\textsubscript{DP}$\downarrow$}
		& \textit{F\textsubscript{OAE}$\downarrow$}
		& AUC$\uparrow$
		& & \\
		\midrule
		
		\multirow{3}{*}{\makecell{CNN\\-based}}
		& Xception $_{\text{\color{blue}CVPR'17}}$\cite{xception}   
		& 39.87 & 37.73 & 36.44 & 22.82 & 50.36
		& 32.36 & 18.66 & 11.75 & 29.71 & 50.44
		& 18.04 & 14.42 & 13.21 & 3.86 & 50.31
		& 55.67 & 37.17 & 33.71 & 15.83 & 42.05
		& 20.81 & 9.19 \\
		
		& ResNet $_{\text{\color{blue}CVPR'16}}$\cite{resnet}   
		& 35.96 & 34.02 & 33.51 & 18.46 & 51.28
		& 13.54 & 8.75 & 6.31 & \textbf{25.48} & 57.12
		& 16.72& 8.64 & 7.59 & 2.89 & 53.24
		& 13.78 & 11.02& 7.91 & 12.04 & 56.99
		& 11.18 & 3.63 \\
		
		& VGG $_{\text{\color{blue}ICLR'15}}$\cite{vgg}   
		& 37.29 & 35.67 & 34.49 & 20.52 & 50.97
		& 15.68& 9.02 & 7.14 & \underline{26.13} & 51.77
		& 17.52& 13.49 & 12.09&3.79 & 50.43
		& \underline{12.50} & \underline{10.36} & \underline{5.49}& \underline{10.85}& \underline{61.83}
		& 134.28 & 30.93 \\
		
		\cmidrule{1-24}
		
		\multirow{3}{*}{\makecell{Trans\\-based}}
		& ViT $_{\text{\color{blue}ICLR'21}}$\cite{ViT}   
		& 39.22 & 25.36 & 23.05 & 13.68 & 51.58
		& 29.80 & 31.08 & 28.68 & 31.89 & 50.93
		& 19.35 & 11.29 & 10.91 & 3.93 & 52.22
		& 44.89 & 36.38 & 30.91 & 23.44 & 33.00
		& 85.77 & 35.14\\
		
		& CViT $_{\text{\color{blue}Arxiv'21}}$\cite{CViT}   
		& 39.99 & 35.42 & 27.02& 25.66 & 49.79
		& 30.36 & 31.34 & 29.28&40.44& 50.28
		& 20.18 &12.15 & 11.53 & 4.75 & 49,89
		& 40.66 &33.46& 30.32 & 16.26&48.47
		& 89.02 & 13.29 \\
		
		& CAEL $_{\text{\color{blue}TIFS'24}}$\cite{genface}   
		& 40.31 & 38.61 & 30.53 & 28.22 & 48.42
		& 35.89 & 34.70 & 33.82 & 40.51& 44.82
		& 22.41 & 15.70 &14.57& 5.58 & 49.02
		& 41.93& 35.74& 30.83 & 19.00 & 42.23
		& 169.38 & 5.01 \\
		
		\cmidrule{1-24}
		
		\multirow{3}{*}{\makecell{VLM\\-based}}
		& CLIP $_{\text{\color{blue}ICML'21}}$\cite{clip}   
		& 22.85 & 21.02 & 20.67 & 13.39 & 49.94
		& 16.71 & 9.05 & 7.38 & 26.36 & 48.15
		& 11.80 & 7.94 & 6.96 & 3.36 & 50.50
		& 51.31 & 39.01 & 27.33 & 27.30 & 32.03
		& 151.28 & 14.69 \\
		
		& MFCLIP $_{\text{\color{blue}TIFS'25}}$\cite{MFCLIP}   
		& 18.58 & 15.77 & 10.48 & 7.71 & 51.04
		& 6.63 & 5.19 & 4.01 & 32.90 & 54.05
		& 5.42 & 4.71 & 3.48 & \textbf{1.06} & 50.85
		& 23.56 & 24.90 & 14.17 & 20.87 & 49.87
		& 141.66 & 39.90 \\
		
		& FA $_{\text{\color{blue}CVPR'25}}$\cite{forensics}       
		& \underline{10.80} & \underline{9.59} & \underline{7.42} & \underline{5.43} & \underline{58.92}
		& 3.76 & 2.93 & 0.86 & 40.34 & \underline{57.15}
		& \underline{4.32} & \underline{3.42} & \underline{3.13} & 2.02 & \underline{55.73}
		& 22.82 & 16.40 & 10.50 & 14.81 & 43.10
		& 435.54 & 348.80 \\
		
		\cmidrule{1-24}
		
		\multirow{7}{*}{\makecell{Fair\\-based}}
		& DAG-FDD $_{\text{\color{blue}CVPR'24}}$ \cite{daw}     
		& 20.53 & 16.60 & 14.40 & 10.50 & 49.96
		& \underline{4.29} & \textbf{2.58} & \textbf{1.04} & 39.83 & 56.81
		& 11.86 & 7.37 & 6.91 & 3.39 & 53.97
		& 21.87 & 22.47 & 11.10 & 28.29 & 41.36
		& 21.86 & 9.10 \\
		
		& DAW-FDD $_{\text{\color{blue}CVPR'24}}$\cite{daw}     
		& 20.99 & 18.85 &15.80 & 12.37& 48.19
		& 10.63 & 7.66 & 6.56 & 40.31 & 49.25
		& 12.82 & 8.28 &7.35 & 4.66 & 50.94
		& 17.09 & 13.48 & 9.90 & 22.62 & 45.67
		& 21.86 & 9.10 \\
		
		& PG-FDD $_{\text{\color{blue}CVPR'24}}$\cite{PFGDD}   
		& 17.59 &10.36& 10.11 & 6.63 & 51.70
		& 9.22 & 7.41 & 6.17 & 40.24& 51.80
		& 11.74 & 8.09 & 7.04 & 4.17 & 51.83
		& 16.47 & 12.93& 9.50 & 16.58& 46.78
		& 69.49 & 28.47 \\
		
		& FairAdapter $_{\text{\color{blue}ICASSP'25}}$\cite{fairadapter}
		& 23.78 & 22.64 & 20.99 & 17.32 & 44.81
		& 18.34 & 10.28& 8.12 & 41.36 &42.69
		& 23.25 & 16.83 & 14.09 & 6.94 & 45.70
		& 46.72& 38.06& 34.75 & 25.80 & 33.82
		& 441.75& 80.69 \\
		
		& RSEF-FDD $_{\text{\color{blue}Arxiv'24}}$\cite{RSEFFDD}
		& 18.94 & 11.56 & 10.84& 8.90 & 50.48
		& 9.80 & 7.53 & 6.48 & 40,29 & 51.73
		& 11.83& 7.99 & 7.01 & 4.03& 52.12
		& 15.94 & 11.76 & 8.48 & 14.67& 51.65
		& 24.05& 9.83\\
		
		& Ding et al. $_{\text{\color{blue}CVPR'26}}$\cite{ding}   
		& 16.79& 10.08& 9.47 &6.13 & 52.13
		& 5.03 & 3.96 & 3.02 & 40.04 & 54.09
		& 8.62 & 5.95 & 4.28 & 2.97 & 54.76
		& 14.58 & 12.09& 8.07 & 13.71 & 54.11
		& 20.81 & 9.91 \\
		
	\rowcolor{cyan!12} & FairForensics (Ours)
		& \textbf{10.45} & \textbf{8.26} &\textbf{6.69} &\textbf{3.53}& \textbf{63.04}
		& \textbf{3.05} &\underline{2.77}& \underline{1.45}& 33.89 & \textbf{62.35}
		& \textbf{3.89} & \textbf{2.94} & \textbf{2.08} & \underline{1.98} & \textbf{59.88}
		& \textbf{10.32}& \textbf{8.55} & \textbf{3.91} & \textbf{8.07} & \textbf{62.75}
		& 421.587& 49.85 \\
				\bottomrule
		\bottomrule
	\end{tabular}
\vspace*{-2em}
\end{table*}

\begin{table}[t]
	\centering
	\caption{Impacts of different language prompts. A, S, G, and R denote authenticity, sex, age, and race prompts, respectively. * and $\dagger$ denote prompts generated with accurate and inaccurate demographic text annotations, respectively.}
	\label{langpro}
	\renewcommand{\arraystretch}{1.35}
	\setlength{\tabcolsep}{0.7pt}
	\scriptsize
	\begin{tabular}{|l|rr|r|rr|r|rr|r|}
		\hline
		\multicolumn{1}{|c|}{\multirow{3}{*}{Models}}
		& \multicolumn{3}{c|}{FF++}
		& \multicolumn{3}{c|}{Celeb-DF}
		& \multicolumn{3}{c|}{DFDC} \\
		\cline{2-10}
		
		& \multicolumn{2}{c|}{Fairness}
		& \multicolumn{1}{c|}{Detect.}
		& \multicolumn{2}{c|}{Fairness}
		& \multicolumn{1}{c|}{Detect.}
		& \multicolumn{2}{c|}{Fairness}
		& \multicolumn{1}{c|}{Detect.} \\
		\cline{2-10}
		
		& \multicolumn{1}{c}{\textit{F\textsubscript{FPR}$\downarrow$}}
		& \multicolumn{1}{c|}{\textit{F\textsubscript{MEO}$\downarrow$}}
		& \multicolumn{1}{c|}{AUC$\uparrow$}
		& \multicolumn{1}{c}{\textit{F\textsubscript{FPR}$\downarrow$}}
		& \multicolumn{1}{c|}{\textit{F\textsubscript{MEO}$\downarrow$}}
		& \multicolumn{1}{c|}{AUC$\uparrow$}
		& \multicolumn{1}{c}{\textit{F\textsubscript{FPR}$\downarrow$}}
		& \multicolumn{1}{c|}{\textit{F\textsubscript{MEO}$\downarrow$}}
		& \multicolumn{1}{c|}{AUC$\uparrow$} \\
		\hline
		
		MFCLIP w/A
		& \multicolumn{1}{c}{18.58} & \multicolumn{1}{c|}{15.77} & \multicolumn{1}{c|}{51.04}
		& \multicolumn{1}{c}{6.63}  & \multicolumn{1}{c|}{5.19}  & \multicolumn{1}{c|}{54.05}
		& \multicolumn{1}{c}{5.42}  & \multicolumn{1}{c|}{4.71}  & \multicolumn{1}{c|}{50.85} \\
		
		MFCLIP w/AS
		& \multicolumn{1}{c}{16.88} & \multicolumn{1}{c|}{11.51}  & \multicolumn{1}{c|}{53.79}
		& \multicolumn{1}{c}{6.78} & \multicolumn{1}{c|}{6.04} & \multicolumn{1}{c|}{53.85}
		& \multicolumn{1}{c}{5.96}  & \multicolumn{1}{c|}{4.39} & \multicolumn{1}{c|}{51.13} \\
		
		MFCLIP w/ASG
		& \multicolumn{1}{c}{14.26} & \multicolumn{1}{c|}{9.68} & \multicolumn{1}{c|}{51.94}
		& \multicolumn{1}{c}{6.65} & \multicolumn{1}{c|}{4.89} & \multicolumn{1}{c|}{54.48}
		& \multicolumn{1}{c}{4.91} & \multicolumn{1}{c|}{3.43} & \multicolumn{1}{c|}{50.57} \\
		
		MFCLIP w/ASGR*
			& \multicolumn{1}{c}{10.91} & \multicolumn{1}{c|}{8.74} & \multicolumn{1}{c|}{54.65}
		& \multicolumn{1}{c}{3.32} & \multicolumn{1}{c|}{2.45} & \multicolumn{1}{c|}{55.44}
		& \multicolumn{1}{c}{3.90} & \multicolumn{1}{c|}{2.96} & \multicolumn{1}{c|}{51.04} \\
	
		MFCLIP w/ASGR†
				& \multicolumn{1}{c}{13.70} & \multicolumn{1}{c|}{10.14}  & \multicolumn{1}{c|}{54.13}
	& \multicolumn{1}{c}{9.27}  & \multicolumn{1}{c|}{6.18}  & \multicolumn{1}{c|}{55.75}
	& \multicolumn{1}{c}{9.63}  & \multicolumn{1}{c|}{8.79} & \multicolumn{1}{c|}{50.47} \\
		
		\hline
		
		Ours w/A
	& \multicolumn{1}{c}{17.21} & \multicolumn{1}{c|}{10.04} & \multicolumn{1}{c|}{59.97}
& \multicolumn{1}{c}{5.08} & \multicolumn{1}{c|}{4.18} & \multicolumn{1}{c|}{61.97}
& \multicolumn{1}{c}{4.25} & \multicolumn{1}{c|}{3.68} & \multicolumn{1}{c|}{59.62} \\

		Ours w/AS
	& \multicolumn{1}{c}{15.53} & \multicolumn{1}{c|}{9.84} & \multicolumn{1}{c|}{61.33}
& \multicolumn{1}{c}{6.71} & \multicolumn{1}{c|}{5.89} & \multicolumn{1}{c|}{61.75}
& \multicolumn{1}{c}{4.67} & \multicolumn{1}{c|}{3.92} & \multicolumn{1}{c|}{59.94} \\

		Ours w/ASA
	& \multicolumn{1}{c}{14.21} & \multicolumn{1}{c|}{8.54} & \multicolumn{1}{c|}{60.16}
& \multicolumn{1}{c}{6.01} & \multicolumn{1}{c|}{4.78} & \multicolumn{1}{c|}{61.66}
& \multicolumn{1}{c}{4.07} & \multicolumn{1}{c|}{3.35} & \multicolumn{1}{c|}{58.93} \\

		\rowcolor{cyan!12}	Ours w/ASAR*
		& \multicolumn{1}{c}{\textbf{10.45}}& \multicolumn{1}{c|}{\textbf{8.26} }& \multicolumn{1}{c|}{\textbf{63.04}}
	& \multicolumn{1}{c}{\textbf{3.05}} & \multicolumn{1}{c|}{\textbf{2.77}}& \multicolumn{1}{c|}{\textbf{62.35}}
	& \multicolumn{1}{c}{\textbf{3.89}} & \multicolumn{1}{c|}{\textbf{2.94}} & \multicolumn{1}{c|}{\textbf{59.88}} \\
		
		Ours w/ASAR†
		& \multicolumn{1}{c}{16.24} & \multicolumn{1}{c|}{9.96} & \multicolumn{1}{c|}{62.87}
	& \multicolumn{1}{c}{8.44} & \multicolumn{1}{c|}{4.09} & \multicolumn{1}{c|}{62.01}
	& \multicolumn{1}{c}{6.88} & \multicolumn{1}{c|}{5.80} & \multicolumn{1}{c|}{59.21} \\
		\hline
	\end{tabular}
\end{table}

\begin{table}[t]
	\centering
	\caption{Effects of loss functions and expression injector.}
	\label{loss}
	\renewcommand{\arraystretch}{1.35}
	\setlength{\tabcolsep}{0.9pt}
	\scriptsize
	\begin{tabular}{|l|rr|r|rr|r|rr|r|}
		\hline
		\multicolumn{1}{|c|}{\multirow{3}{*}{\multirow{2}{*}{\makecell{Loss\\Function}}}}
		& \multicolumn{3}{c|}{FF++}
		& \multicolumn{3}{c|}{Celeb-DF}
		& \multicolumn{3}{c|}{DFDC} \\
		\cline{2-10}
		
		& \multicolumn{2}{c|}{Fairness}
		& \multicolumn{1}{c|}{Detect.}
		& \multicolumn{2}{c|}{Fairness}
		& \multicolumn{1}{c|}{Detect.}
		& \multicolumn{2}{c|}{Fairness}
		& \multicolumn{1}{c|}{Detect.} \\
		\cline{2-10}
		
		& \multicolumn{1}{c}{\textit{F\textsubscript{FPR}$\downarrow$}}
		& \multicolumn{1}{c|}{\textit{F\textsubscript{MEO}$\downarrow$}}
		& \multicolumn{1}{c|}{AUC$\uparrow$}
		& \multicolumn{1}{c}{\textit{F\textsubscript{FPR}$\downarrow$}}
		& \multicolumn{1}{c|}{\textit{F\textsubscript{MEO}$\downarrow$}}
		& \multicolumn{1}{c|}{AUC$\uparrow$}
		& \multicolumn{1}{c}{\textit{F\textsubscript{FPR}$\downarrow$}}
		& \multicolumn{1}{c|}{\textit{F\textsubscript{MEO}$\downarrow$}}
		& \multicolumn{1}{c|}{AUC$\uparrow$} \\
		\hline

			Ours w/o $\mathcal{L}_{\mathrm{efd}}$
		& \multicolumn{1}{c}{14.05} & \multicolumn{1}{c|}{12.43} & \multicolumn{1}{c|}{60.77}
		& \multicolumn{1}{c}{6.35} & \multicolumn{1}{c|}{5.89} & \multicolumn{1}{c|}{60.35}
		& \multicolumn{1}{c}{6.06} & \multicolumn{1}{c|}{5.90} & \multicolumn{1}{c|}{55.41} \\
		 			Ours w/o $\mathcal{L}_{\mathrm{vfa}}$
		& \multicolumn{1}{c}{13.98} & \multicolumn{1}{c|}{11.04} & \multicolumn{1}{c|}{61.83}
		& \multicolumn{1}{c}{5.60} & \multicolumn{1}{c|}{4.71} & \multicolumn{1}{c|}{60.94}
		& \multicolumn{1}{c}{5.37} & \multicolumn{1}{c|}{4.82} & \multicolumn{1}{c|}{56.09} \\
			Ours w/o $\mathcal{L}_{\mathrm{ppf}}$
		& \multicolumn{1}{c}{18.37} & \multicolumn{1}{c|}{13.74} & \multicolumn{1}{c|}{58.65}
		& \multicolumn{1}{c}{8.93} & \multicolumn{1}{c|}{7.02} & \multicolumn{1}{c|}{57.41}
		& \multicolumn{1}{c}{9.04} & \multicolumn{1}{c|}{8.31} & \multicolumn{1}{c|}{52.36} \\
			Ours w/o $\mathcal{L}_{\mathrm{dsad}}$
		& \multicolumn{1}{c}{11.28} & \multicolumn{1}{c|}{9.74} & \multicolumn{1}{c|}{62.24}
		& \multicolumn{1}{c}{4.90} & \multicolumn{1}{c|}{3.04} & \multicolumn{1}{c|}{61.02}
		& \multicolumn{1}{c}{4.73} & \multicolumn{1}{c|}{3.69} & \multicolumn{1}{c|}{58.62} \\
					Ours w/o $\mathcal{L}_{\mathrm{mar}}$
		& \multicolumn{1}{c}{11.92} & \multicolumn{1}{c|}{9.84} & \multicolumn{1}{c|}{61.76}
		& \multicolumn{1}{c}{6.47} & \multicolumn{1}{c|}{5.36} &
		\multicolumn{1}{c|}{60.42}
		& \multicolumn{1}{c}{6.08} & \multicolumn{1}{c|}{4.87} & \multicolumn{1}{c|}{58.42} \\
		
		Ours w/o $\mathcal{L}_{\mathrm{align}}$
		& \multicolumn{1}{c}{14.28} & \multicolumn{1}{c|}{11.63} & \multicolumn{1}{c|}{60.98}
		& \multicolumn{1}{c}{4.96} & \multicolumn{1}{c|}{4.10} &
		\multicolumn{1}{c|}{60.93} 
		& \multicolumn{1}{c}{7.95} & \multicolumn{1}{c|}{6.71} & \multicolumn{1}{c|}{57.73} \\
			Ours w/o EI
		& \multicolumn{1}{c}{11.34} & \multicolumn{1}{c|}{9.90} & \multicolumn{1}{c|}{60.87}
		& \multicolumn{1}{c}{4.75} & \multicolumn{1}{c|}{3.83} & \multicolumn{1}{c|}{60.26}
		& \multicolumn{1}{c}{4.07} & \multicolumn{1}{c|}{3.39} & \multicolumn{1}{c|}{58.92} \\
					\rowcolor{cyan!12} 			FairForensics (ours)
		& \multicolumn{1}{c}{\textbf{10.45}} & \multicolumn{1}{c|}{\textbf{8.26}} & \multicolumn{1}{c|}{\textbf{63.04}}
		& \multicolumn{1}{c}{\textbf{3.05}} & \multicolumn{1}{c|}{\textbf{2.77}}& \multicolumn{1}{c|}{\textbf{62.35}}
		& \multicolumn{1}{c}{\textbf{3.89}} & \multicolumn{1}{c|}{\textbf{2.94}} & \multicolumn{1}{c|}{\textbf{59.88}} \\
		\hline
		
	\end{tabular}
\vspace*{-2em}
\end{table}

\begin{table}[t]
	\centering
	\caption{Impacts of different expression estimators. * † ‡ denote the SSFER pretrained using the RAFDB, FERPLUS, and AffectNet8 datasets, respectively.}
	\label{expes}
	\renewcommand{\arraystretch}{1.35}
	\setlength{\tabcolsep}{2.0pt}
	\scriptsize
	\begin{tabular}{|l|rr|r|rr|r|rr|r|}
		\hline
		\multicolumn{1}{|c|}{\multirow{3}{*}{\multirow{2}{*}{\makecell{Expression\\Estimator}}}}
		& \multicolumn{3}{c|}{FF++}
		& \multicolumn{3}{c|}{Celeb-DF}
		& \multicolumn{3}{c|}{DFDC} \\
		\cline{2-10}
		
		& \multicolumn{2}{c|}{Fairness}
		& \multicolumn{1}{c|}{Detect.}
		& \multicolumn{2}{c|}{Fairness}
		& \multicolumn{1}{c|}{Detect.}
		& \multicolumn{2}{c|}{Fairness}
		& \multicolumn{1}{c|}{Detect.} \\
		\cline{2-10}
		
		& \multicolumn{1}{c}{\textit{F\textsubscript{FPR}$\downarrow$}}
		& \multicolumn{1}{c|}{\textit{F\textsubscript{MEO}$\downarrow$}}
		& \multicolumn{1}{c|}{AUC$\uparrow$}
		& \multicolumn{1}{c}{\textit{F\textsubscript{FPR}$\downarrow$}}
		& \multicolumn{1}{c|}{\textit{F\textsubscript{MEO}$\downarrow$}}
		& \multicolumn{1}{c|}{AUC$\uparrow$}
		& \multicolumn{1}{c}{\textit{F\textsubscript{FPR}$\downarrow$}}
		& \multicolumn{1}{c|}{\textit{F\textsubscript{MEO}$\downarrow$}}
		& \multicolumn{1}{c|}{AUC$\uparrow$} \\
		\hline

			SSFER*
		& \multicolumn{1}{c}{11.14} & \multicolumn{1}{c|}{9.73} & \multicolumn{1}{c|}{53.99}
		& \multicolumn{1}{c}{17.41} & \multicolumn{1}{c|}{7.41} & \multicolumn{1}{c|}{53.14}
		& \multicolumn{1}{c}{5.02} & \multicolumn{1}{c|}{3.66} & \multicolumn{1}{c|}{50.92} \\
		
			SSFER†
		& \multicolumn{1}{c}{14.65} & \multicolumn{1}{c|}{11.38} & \multicolumn{1}{c|}{51.60}
		& \multicolumn{1}{c}{25.59} & \multicolumn{1}{c|}{8.93} & \multicolumn{1}{c|}{46.95}
		& \multicolumn{1}{c}{5.01} & \multicolumn{1}{c|}{\textbf{1.30}} & \multicolumn{1}{c|}{51.59} \\
		
		SSFER‡
		& \multicolumn{1}{c}{15.03} & \multicolumn{1}{c|}{10.47}  & \multicolumn{1}{c|}{50.26}
		& \multicolumn{1}{c}{25.81}  & \multicolumn{1}{c|}{9.03}  & \multicolumn{1}{c|}{44.66}
		& \multicolumn{1}{c}{6.13}  & \multicolumn{1}{c|}{5.84} & \multicolumn{1}{c|}{50.31} \\
		
			\rowcolor{cyan!12} 	POSTER
		& \multicolumn{1}{c}{\textbf{10.45}} & \multicolumn{1}{c|}{\textbf{8.26}} & \multicolumn{1}{c|}{\textbf{63.04}}
		& \multicolumn{1}{c}{\textbf{3.05}} & \multicolumn{1}{c|}{\textbf{2.77}} & \multicolumn{1}{c|}{\textbf{62.35}}
		& \multicolumn{1}{c}{\textbf{3.89}} & \multicolumn{1}{c|}{2.94} & \multicolumn{1}{c|}{\textbf{59.88}} \\
		\hline
	
	\end{tabular}
\vspace*{-1em}
\end{table}

\begin{table}[t]
	\centering
	\caption{Ablations of the plug-and-play IAPP.}
	\label{iapp}
	\renewcommand{\arraystretch}{1.35}
	\setlength{\tabcolsep}{0.2pt}
	\scriptsize
	\begin{tabular}{|l|rr|r|rr|r|rr|r|rr|}
		\hline
		\multicolumn{1}{|c|}{\multirow{3}{*}{\makecell{IAPP}}}
		& \multicolumn{3}{c|}{FF++}
		& \multicolumn{3}{c|}{Celeb-DF}
		& \multicolumn{3}{c|}{DFDC}
		& \multicolumn{2}{c|}{\multirow{2}{*}{\makecell{Complexity\\Cost}}} \\
		\cline{2-10}
		
		& \multicolumn{2}{c|}{Fairness}
		& \multicolumn{1}{c|}{Detect.}
		& \multicolumn{2}{c|}{Fairness}
		& \multicolumn{1}{c|}{Detect.}
		& \multicolumn{2}{c|}{Fairness}
		& \multicolumn{1}{c|}{Detect.}
		& \multicolumn{2}{c|}{} \\
		\cline{2-12}
		
		& \multicolumn{1}{c}{\textit{F\textsubscript{FPR}$\downarrow$}}
		& \multicolumn{1}{c|}{\textit{F\textsubscript{MEO}$\downarrow$}}
		& \multicolumn{1}{c|}{AUC$\uparrow$}
		& \multicolumn{1}{c}{\textit{F\textsubscript{FPR}$\downarrow$}}
		& \multicolumn{1}{c|}{\textit{F\textsubscript{MEO}$\downarrow$}}
		& \multicolumn{1}{c|}{AUC$\uparrow$}
		& \multicolumn{1}{c}{\textit{F\textsubscript{FPR}$\downarrow$}}
		& \multicolumn{1}{c|}{\textit{F\textsubscript{MEO}$\downarrow$}}
		& \multicolumn{1}{c|}{AUC$\uparrow$}
		& \multicolumn{1}{c}{Params}
		& \multicolumn{1}{c|}{FLOPs} \\
		\hline
		
		ViT w/o
		& \multicolumn{1}{c}{39.22} & \multicolumn{1}{c|}{25.36} & \multicolumn{1}{c|}{51.58}
		& \multicolumn{1}{c}{29.80} & \multicolumn{1}{c|}{31.08} & \multicolumn{1}{c|}{50.93}
		& \multicolumn{1}{c}{19.35} & \multicolumn{1}{c|}{11.29} & \multicolumn{1}{c|}{52.22}
		& \multicolumn{1}{c}{85.77} & \multicolumn{1}{c|}{35.14} \\
		
			ViT w/
		& \multicolumn{1}{c}{38.20} & \multicolumn{1}{c|}{24.19} & \multicolumn{1}{c|}{51.72}
		& \multicolumn{1}{c}{22.65} & \multicolumn{1}{c|}{27.89} & \multicolumn{1}{c|}{51.06}
		& \multicolumn{1}{c}{13.93} & \multicolumn{1}{c|}{10.12} & \multicolumn{1}{c|}{52.29}
		& \multicolumn{1}{c}{89.02} & \multicolumn{1}{c|}{35.25} \\
		
	Ours w/o
		& \multicolumn{1}{c}{10.56} & \multicolumn{1}{c|}{9.89}  & \multicolumn{1}{c|}{60.22}
		& \multicolumn{1}{c}{10.43}  & \multicolumn{1}{c|}{6.05}  & \multicolumn{1}{c|}{59.46}
		& \multicolumn{1}{c}{9.32}  & \multicolumn{1}{c|}{4.79} & \multicolumn{1}{c|}{58.09}
		& \multicolumn{1}{c}{421.29} & \multicolumn{1}{c|}{49.83} \\
		
		\rowcolor{cyan!12} 		Ours w/
		& \multicolumn{1}{c}{\textbf{10.45}} & \multicolumn{1}{c|}{\textbf{8.26}} & \multicolumn{1}{c|}{\textbf{63.04}}
		& \multicolumn{1}{c}{\textbf{3.05}} & \multicolumn{1}{c|}{\textbf{2.77}}& \multicolumn{1}{c|}{\textbf{62.35}}
		& \multicolumn{1}{c}{\textbf{3.89}}& \multicolumn{1}{c|}{\textbf{2.94}} & \multicolumn{1}{c|}{\textbf{59.88}}
		& \multicolumn{1}{c}{421.59} & \multicolumn{1}{c|}{49.85} \\
		\hline
		
	\end{tabular}
\vspace*{-2em}
\end{table}

\begin{table}[t]
	\centering
	\caption{Model ablation. Fairness and detection scores (\%) on FF++, Celeb-DF, and DFDC after training using the balanced demographic face data in GenFace.}
	\label{modelab}
	\renewcommand{\arraystretch}{1.35}
	\setlength{\tabcolsep}{0.3pt}
	\scriptsize
	\begin{tabular}{|c|c|c|rr|r|rr|r|rr|r|r|}
		\hline
		
		\multicolumn{3}{|c|}{\multirow{3}{*}{Models}}
		& \multicolumn{3}{c|}{FF++}
		& \multicolumn{3}{c|}{Celeb-DF}
		& \multicolumn{3}{c|}{DFDC}
		& \multicolumn{1}{c|}{Avg.} \\
		
		\cline{4-13}
		
		\multicolumn{3}{|c|}{}
		& \multicolumn{2}{c|}{Fairness}
		& \multicolumn{1}{c|}{Detect.}
		& \multicolumn{2}{c|}{Fairness}
		& \multicolumn{1}{c|}{Detect.}
		& \multicolumn{2}{c|}{Fairness}
		& \multicolumn{1}{c|}{Detect.}
		& \multicolumn{1}{c|}{Detect.} \\
		
		\cline{4-13}
		
		\cline{1-3}
		
		EE & EPVE & DGLE
		& \multicolumn{1}{c}{\textit{F\textsubscript{FPR}$\downarrow$}}
		& \multicolumn{1}{c|}{\textit{F\textsubscript{MEO}$\downarrow$}}
		& \multicolumn{1}{c|}{AUC$\uparrow$}
		& \multicolumn{1}{c}{\textit{F\textsubscript{FPR}$\downarrow$}}
		& \multicolumn{1}{c|}{\textit{F\textsubscript{MEO}$\downarrow$}}
		& \multicolumn{1}{c|}{AUC$\uparrow$}
		& \multicolumn{1}{c}{\textit{F\textsubscript{FPR}$\downarrow$}}
		& \multicolumn{1}{c|}{\textit{F\textsubscript{MEO}$\downarrow$}}
		& \multicolumn{1}{c|}{AUC$\uparrow$}
		& \multicolumn{1}{c|}{AUC$\uparrow$}\\
		
		\hline
		
		$\checkmark$ & & 
		& 18.19 & 16.40 & 54.43
		& 15.80 & 10.47 & 50.19
		& 10.12 & 7.91 & 48.69
		& 51.10 \\
		
		& $\checkmark$ & 
		& 16.02 & 15.88 & 56.37
		& 14.62 & 9.60 & 53.64
		& 8.89 & 6.21 & 50.69
		& 53.57 \\
		
		$\checkmark$ & $\checkmark$ & 
		& 17.97 & 14.15 & 58.99
		& 11.87 & 6.40 & 55.72
		& 6.91 & 5.03 & 53.52
		& 56.08 \\
		
		$\checkmark$ & & $\checkmark$
		& 14.78 & 13.34 & 56.13
		& 10.27 & 6.18 & 52.75
		& 6.63 & 5.79 & 50.45
		& 53.11 \\
		
		& $\checkmark$ & $\checkmark$
		& 13.47 & 11.08 & 59.46
		& 5.89 & 4.05 & 58.87
		& 4.26 & 3.88 & 56.39
		& 58.24 \\
		
		\rowcolor{cyan!12}
		$\checkmark$ & $\checkmark$ & $\checkmark$
		& \textbf{10.45} & \textbf{8.26} & \textbf{63.04}
		& \textbf{3.05} & \textbf{2.77} & \textbf{62.35}
		& \textbf{3.89} & \textbf{2.94} & \textbf{59.88}
		& \textbf{61.76} \\
		
		\hline
		
	\end{tabular}
\vspace*{-2em}
\end{table}

\begin{figure*}[t!]
	\centering
	\includegraphics[width=\linewidth]{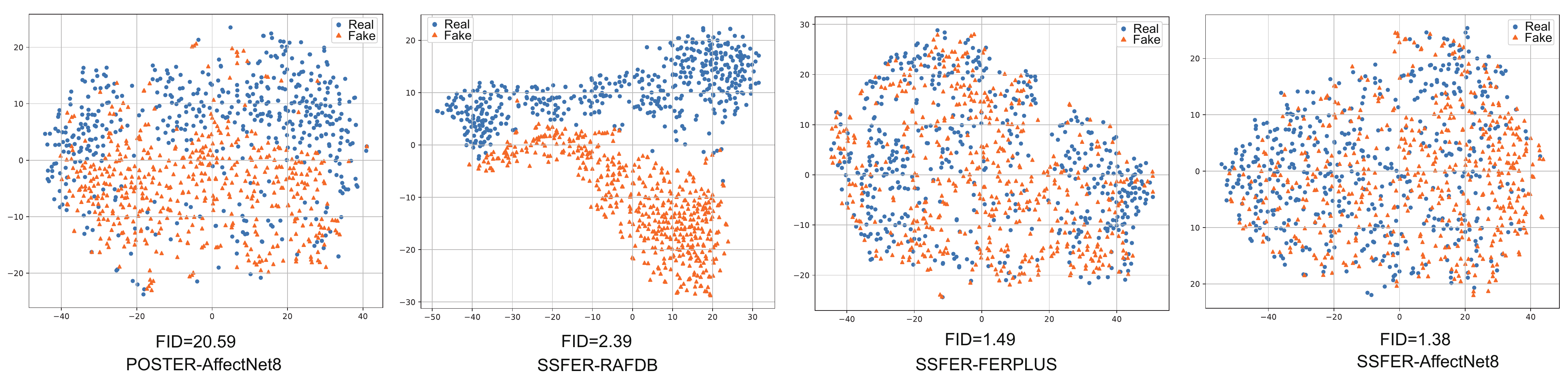}
	\caption{t-SNE visualization of expression embeddings extracted by POSTER pretrained using AffectNet8 and SSFER pretrained using RAFDB, FERPLUS, and AffectNet8. We randomly select 10k real expression embeddings and 10k fake ones to calculate the FID. The higher the FID score, the greater the distribution difference between the real and fake expression embeddings.}\label{expvisual}
	\vspace{-2em}
\end{figure*}

{\bfseries\setlength\parindent{0em} Within-dataset evaluation.} We investigate the influence of balanced and unbalanced intersectional demographic subgroup (IDS) settings on detector fairness. The total number of samples in the imbalanced setting is kept consistent with that of the balanced one, while the distribution follows the original imbalanced distribution of GenFace. Since FDD is a relatively new and challenging task, to the best of our knowledge, prior work has hardly evaluated the fairness of FDD models under both balanced and imbalanced settings systematically and comprehensively. Therefore, we present the first systematic assessment of model fairness across these scenarios. Specifically, models are trained on either balanced or imbalanced IDS from GenFace \cite{genface} and evaluated on both testing settings. As shown in Table~\ref{within}, models trained using balanced IDS distributions generally achieve lower fairness scores on balanced and unbalanced testing sets than those trained on unbalanced one, demonstrating that balanced demographic datasets can alleviate demographic bias of detectors. This improvement is evident when comparing the performance of the same detector under different training settings. For instance, the $\textit{F\textsubscript{FPR}}$ of FA is 0.21\% and 0.17\% on balanced and unbalanced IDS testing sets after training using balanced IDS, respectively. By contrast, It only acquire 0.43\% and 0.25\% $\textit{F\textsubscript{FPR}}$ on balanced and unbalanced testing sets after training using unbalanced IDS, individually.
This indicates that balanced demographic distributions push the model to learn more equitable decision boundaries across different subgroups. Moreover, balanced IDS datasets also improves detector fairness under demographic distribution shifts. Specifically, models trained with balanced IDS distributions usually exhibit lower fairness scores on the unbalanced IDS testing sets, compared with those trained on unbalanced IDS ones. For example, MFCLIP trained on balanced and unbalanced IDS datasets achieves $\textit{F\textsubscript{FPR}}$ scores of 0.84\% and 0.92\%, respectively, on the unbalanced IDS testing set. This suggests that demographiclly-balanced deepfake dataset can prevent the detector from overfitting to majority groups and enhance its generalization to minority ones. Compared with existing detectors, our method achieves lower fairness scores in most cases while maintaining a higher AUC, indicating that it benefits from balanced IDS dataset training and strengthens fairness through fairness-aware representation learning.

{\bfseries\setlength\parindent{0em} Cross-dataset protocol.}
To provide stronger evidence of domain generalization and demographic fairness beyond the proposed balanced IDS benchmarks, we train networks using our balanced IDS training sets and test them on unseen deepfake datasets, including FF++ \cite{wilddeepfake}, Celeb-DF \cite{Celeb-DF}, DFDC \cite{DFDC}, and DF-1.0 \cite{DF1.0}. As shown in Table~\ref{cross}, our method achieves the best detection performance across all testing datasets among all detectors, with the AUC score of 63.04\%, 62.35\%, 59.88\%, and 62.75\% on FF++, Celeb-DF, DFDC, and DF-1.0, respectively. Compared with CNN, transformer, and VLM-based detectors, our model shows stronger generalization performance on unseen deepfake datasets. We argue that high-level EFP and fairness-aware vision-language learning boost the transferability of our model. Our method also achieves lower fairness gaps on cross-dataset than existing FDD methods. For example, the $F_\text{FPR}$, $F_\text{MEO}$, $F_\text{DP}$, and $F_\text{OAE}$ of our model are 3.89\%, 2.94\%, 2.08\%, and 1.98\% on DFDC, respectively. By contrast, Ding et al. obtain 8.62\% $F_\text{FPR}$, 5.95\% $F_\text{MEO}$, 4.28\% $F_\text{DP}$, and 2.97\% $F_\text{OAE}$, respectively. To conduct the computational efficiency benchmark, we evaluate the parameter counts and FLOPs of detectors. Despite of more parameters than lightweight Xception, FairForensics shows significant improvements in both detection generalization and fairness performance. In particular, the AUC of FairForensics is about 12\% higher than that of Xception on FF++, and the $F_\text{FPR}$ of FairForensics is about 29\% lower than that of Xception, with an increase of about 400M parameters and 40G FLOPs. We argue that our method learns forgery-discriminative yet demographic-invariant representations, thus realizing better detection generalization and demographic fairness under cross-dataset evaluation. 
%\begin{figure}[t!]
%	\centering
%	\includegraphics[width=\linewidth]{Fig4heatmap2.eps}
%	\vspace{-2em}
%	\caption{The heatmap visualizations of our GazeCLIP model (w/o or w/ gaze) on different generator samples. }
%			\vspace{-2em}
%	\label{heat}
%\end{figure}

{\bfseries\setlength\parindent{0em} Robustness analysis.} We evaluate the robustness of different models against unseen image distortions from the DF-1.0 dataset \cite{DF1.0}. As shown in Fig.~\ref{robust2}, ten types of corruptions are considered, including saturation, contrast, block-wise distortion, Gaussian noise, video compression, Gaussian blur, pixelation, upscaling, downscaling, and cropping. Each corruption contains five severities, where severity 0 denotes original clean images. Overall, existing methods show obvious performance degradation under severe perturbations, especially for Gaussian noise, Gaussian blur, video compression, and cropping, indicating that they are sensitive to image quality degradation. In contrast, our method consistently achieves the highest AUC across most corruption types and severity levels. Even under strong distortions, our method maintains relatively stable performance, demonstrating superior robustness to unseen image corruptions. We argue that our model benefits from high-level EFP and demographic-guided vision-language learning to capture more semantically rich and generalizable forgery representations, rather than relying solely on fragile low-level texture artifacts.

%We train networks using our protocol and test their attribution performance on unseen distorted images from \cite{df10}. In Figure~\ref{robust2}, ten types of corruption are engaged, each with five severities. A severity of 0 means no corruption. In detail, specific settings for the intensity of ten deformations are discussed in \cite{genface}. Models are inclined to show sensitivity to deformations like video compression. Our approach significantly outperforms most models across various types of image perturbations. Different from existing methods that focus on low-level texture features that are highly susceptible to image quality degradation, our model considers high-level gaze features related to the consistency of facial pose, eye movement, and facial expression, which makes our method equip strong robustness to image quality variations. 
\vspace{-1em}
\subsection{Ablation Study}
%{\bfseries\setlength\parindent{0em} Impacts of various modules.}
%In Table~\ref{var}, GE increases the attribution performance by about 10\% ACC, showing that gaze features offer valuable clues to boost DFAD. AGPM enhances the ACC by about 16\%, which validates that appearance-gaze global manipulated embeddings are vital for DFAD. The gain from introducing GIE (+5.96\%) is obvious, emphasizing the importance of general gaze-aware image forgery embeddings. LRE improves performance by about 18.63\% ACC when GIE is also involved. The gain brought by LE alone (+5.78\%) is marginal without the introduction of GIE, as the effectiveness of LE depends on the synergy between the various common forgery features. LRE offers fine-grained adaptive-enhanced semantic guidance, which is enhanced when combined with the distinctive gaze prior and general facial image forgery features from GIE. Without the module, LRE's potential is not fully realized, leading to a smaller performance gain.

{\bfseries\setlength\parindent{0em} Impacts of various modules.} We conduct the ablation experiments by training detectors using our balanced IDS datasets and testing them on FF++, CelebDF, and DFDC. In Table~\ref{modelab}, EE enhances the average AUC from 53.57\% to 56.08\%, which validates that explicit expression estimation is beneficial for learning expression-aware forgery representations. EPVE increases the average detection performance from 51.10\% to 56.08\% AUC, showing that expression-perceptual visual forgery features offer valuable clues. The gain from introducing DGLE is obvious, and it improves the average AUC and $\textit{F\textsubscript{FPR}}$ by 5.68\% and 6.45\%, respectively, emphasizing the importance of demographic-guided language embeddings for generalizable FDD. Our model achieves the highest average AUC of 61.76\%, surpassing the combinations of EE with EPVE, EE with DGLE, and EPVE with DGLE by 5.69\%, 8.65\%, and 3.52\%, respectively. Meanwhile, our model also obtains the lowest average fairness scores, reducing $\textit{F\textsubscript{FPR}}$ and $\textit{F\textsubscript{MEO}}$ to 5.80\% and 4.66\%, respectively. This indicates that EE, EPVE, and DGLE are complementary. Without DGLE, our model mainly focuses on facial expression-related forgery cues, and their fairness enhancement is limited. As DGLE is involved, population-aware semantic guidance adaptively regularizes visual representations, yielding more general and fairer forgery features.

{\bfseries\setlength\parindent{0em} Influences of expression estimator.} We delve into the impact of various estimators, such as POSTER pretrained by AffectNet8 \cite{ETHXGaze}, SSFER \cite{ssfer} pretrained using RAFDB, FERPLUS, and AffectNet8 datasets. In Table~\ref{expes}, the detection and fairness performance of our model achieves the maximum when the POSTER is engaged. To provide a clearer justification for the selection of the expression estimators, we further provide a feature distribution analysis using t-SNE, and compute the FID scores by picking 500 expression vectors extracted by various expression estimators from fake and real images. A higher FID score indicates a larger distribution discrepancy between real and fake expression embeddings. In Figure~\ref{expvisual}, the FID value of expression features extracted by POSTER is higher than that of those captured by other SSFER methods, showing that POSTER struggles to generalize to unseen fake faces. A lower FID score does not necessarily imply better performance for FDD. The goal of our expression branch is not to perform conventional face expression recognition, but to offer expression-related cues that are helpful for distinguishing real and fake faces. From this perspective, the relatively larger distribution discrepancy captured by POSTER may preserve more forgery-sensitive expression variations, especially those captured by imperfect expression recognition. These cues can be highly beneficial for FDD, because fake faces often appear visually realistic while still containing subtle expression-level inconsistencies. This observation is also supported by the results in Table~\ref{expes}, where models trained with larger expression embedding distribution differences tend to achieve better fairness and generalization performance on unseen deepfake datasets. We argue that SSFER tends to learn more domain-invariant and recognition-oriented expression features, which are effective for facial expression recognition but may suppress subtle forgery-related expression discrepancies. By contrast, POSTER retains more domain-sensitive expression variations, allowing the detector to effectively exploit EFP patterns.

{\bfseries\setlength\parindent{0em} Effects of identity-aware patch perturbations.}
To evaluate the effectiveness of IAPP, we conduct ablation experiments by integrating it into transformer-based models. As shown in Table~\ref{iapp}, IAPP consistently improves the generalization and fairness performance of various transformer-based methods. Specifically,  due to the introduction of IAPP, the AUC of ViT and FairForensics is improved by 0.2\% and 2.89\% on Celeb-DF after training using our balanced  IDS distributions, with a growth of 0.11M and 2.15M parameters, accordingly. We further investigate the impact of IAPP on fairness performance. We notice that the $\textit{F\textsubscript{FPR}}$ score of ViT and FairForensics is reduced by 7.15\% and 7.38\%, respectively, on Celeb-DF after applying IAPP. We argue that identity-sensitive patch perturbations can effectively alleviate the model's reliance on identity-related shortcut cues, thereby improving demographic fairness under cross-dataset evaluation. Moreover, IAPP introduces only negligible additional parameters and computational overhead, since it uses a lightweight identity scoring head and performs perturbation only during training. We argue that IAPP encourages the model to attend to more diverse and forgery-relevant facial regions instead of overfitting to dominant identity-sensitive patches, leading to fairer and more generalizable forgery representations.
\begin{figure}[t]
	\centering
	\includegraphics[width=\linewidth]{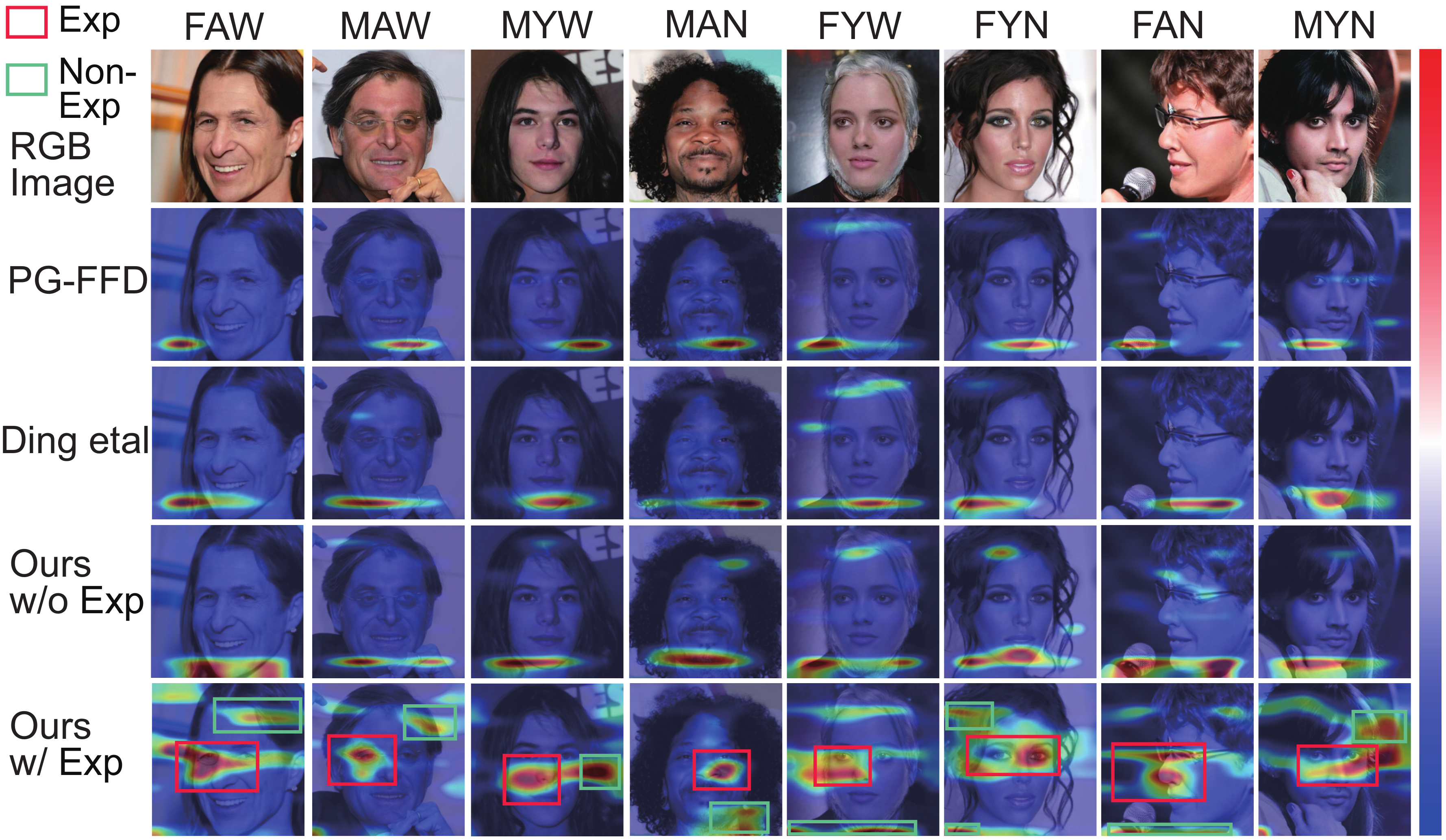}
	\caption{Heatmap visualizations of fairness-enhanced detectors and our model w/o or /with EFP across different demographic subgroups. Hotter colors denote stronger model attention to potential forgery evidence.}
	\vspace*{-1em}
	\label{heatmap}
\end{figure}

%\begin{figure}[t!]%\citep{wodajo2021deepfake,coccomini2022combining}
%	\centering
%	\includegraphics[width=\linewidth]{Demontextrvisual.eps}	
%	\caption{The typical authenticity and demographic levels text prompt examples for various intersectional population groups. }\label{de}
%\end{figure}

{\bfseries\setlength\parindent{0em} Impacts of demographic-aware text prompts.}
	To investigate the effect of demographic-aware text prompts, we evaluate the performance of MFCLIP and our method on the cross-dataset protocol by progressively incorporating different levels of textual prompts. As shown in Table~\ref{langpro}, the performance of our model and MFCLIP commonly improves with the increase of demographic-aware text prompts, demonstrating the importance of explicit demographic semantic guidance for generalizable FDD. Specifically, the AUC of MFCLIP and FairForensics is improved, and the $\textit{F\textsubscript{FPR}}$ is decreased on FF++, Celeb-DF, and DFDC, respectively, as sex, age, and race prompts are introduced, showing that demographic-aware language information helps reduce subgroup bias. When complete and accurate demographic-aware prompts are incorporated, our model achieves the best trade-off between generalization and fairness, with the highest AUC scores and the lowest fairness gaps on most metrics. However, as inaccurate demographic prompts, i.e., a photo of a face of a white race or a photo of a face with a non-white race, are introduced, the fairness scores of both MFCLIP and our method improve noticeably. We argue that accurate demographic-aware prompts guide the model to disentangle manipulation-relevant cues from demographic-related attributes, thereby improving both detection generalization and fairness.

\begin{figure}[t]
	\centering
	\includegraphics[width=\linewidth]{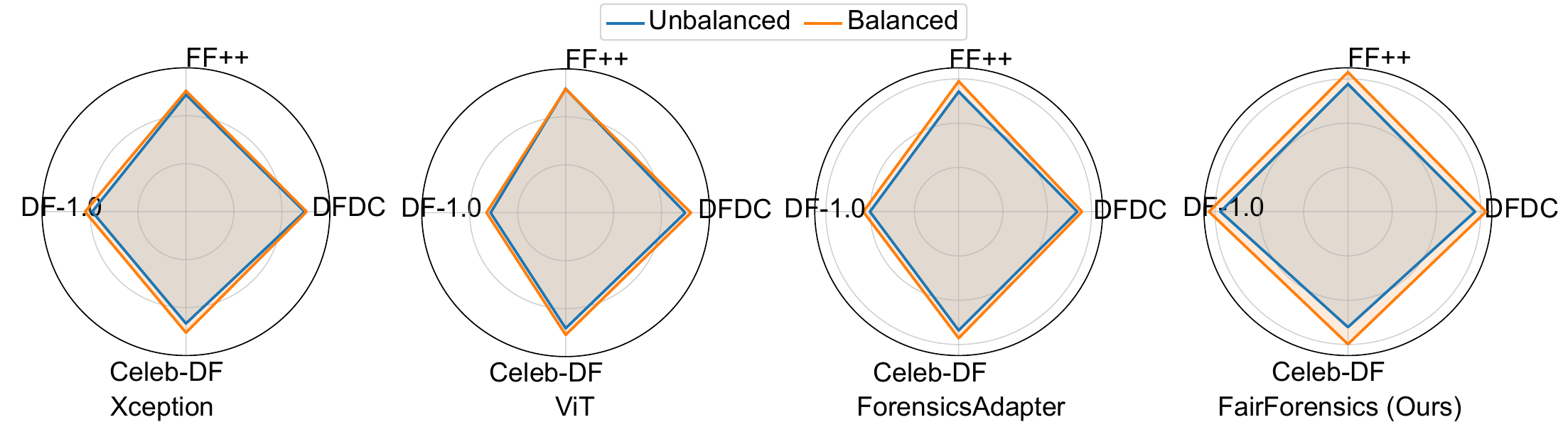}
	\caption{The detection performance of various detectors trained using unbalanced or balanced IDS from GenFace on FF++, DFDC, Celeb-DF, and DF-1.0. } \vspace*{-1em}
	\label{dfadfd}
\end{figure}

\begin{figure*}[t!]%\citep{wodajo2021deepfake,coccomini2022combining}
	\centering
	\includegraphics[width=\linewidth]{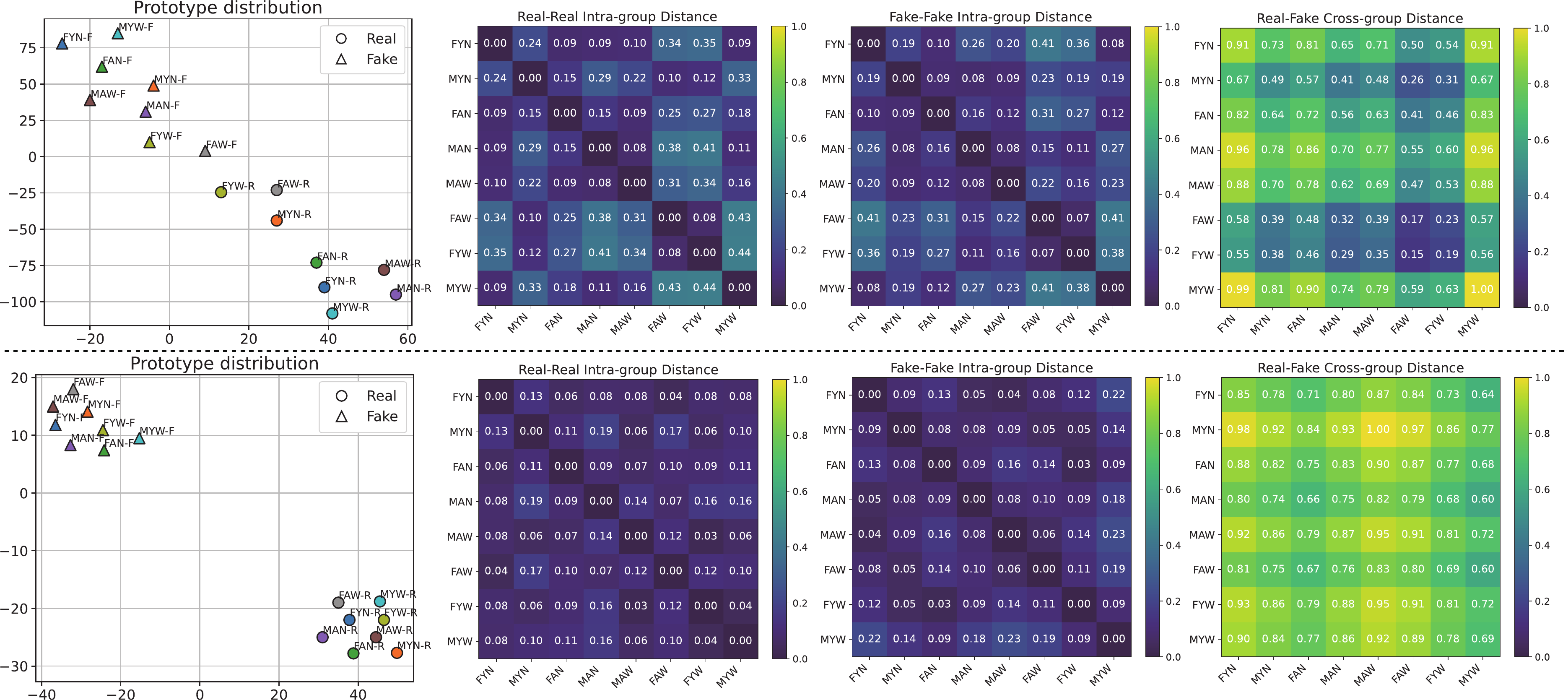} 
	\caption{ Visualization of prototype distribution and intra-/cross-group distance heatmaps of prototypes generated by our model w/o (top) and w/ (bottom) the PPF loss. For each IDS, we randomly select 400 real and 400 fake samples to generate prototypes. The brighter the color, the greater the distance.}\label{prototype}
	\vspace*{-1em}
\end{figure*}

{\bfseries\setlength\parindent{0em} Impacts of balanced intersection groups.} We investigate the effect of balanced IDS on cross-dataset generalization. Specifically, we train different detectors using either unbalanced or balanced IDS from GenFace, and evaluate their detection performance on FF++, DFDC, Celeb-DF, and DF-1.0. As shown in Fig.~\ref{dfadfd}, models trained with balanced demographic distributions generally achieve better detection performance across different testing datasets. We argue that balanced demographic distributions mitigate the dominance of majority groups, which encourages the model to capture more generalizable forgery representations rather than demographic-specific shortcut cues. This improvement is important under cross-dataset evaluation, where the demographic distributions of the testing data may differ from those of the training data. This demonstrates that balanced IDS datasets enhance the transferability of deepfake detectors, leading to better generalization across datasets.

%{\bfseries\setlength\parindent{0em} Effects of the placement of IAPP.} To explore IAPP in greater depth, we further investigate the influence of the placement of IAPP. As shown in Table~\ref{plaiapp}, the performance of our model attains the maximum when IAPP is placed after the first EPTB. We believe that our method with the IAPP placed before the first EPTB applies perturbations to relatively shallow visual tokens that have not yet been sufficiently enriched with EFP, thereby limiting its ability to identify identity-sensitive shortcut regions. In contrast, FairForensics with the IAPP inserted too late (e.g., after the second EPTB) may perturb features that already encode more task-discriminative forgery semantics, leading to degraded detection performance and unstable fairness gains. This suggests that our model with an early but not initial placement of identity-aware perturbation strikes a better balance, effectively suppressing identity-biased patch cues while preserving sufficient manipulation-relevant information for subsequent expression-guided forgery representation learning.

{\bfseries\setlength\parindent{0em} Impacts of various losses.} As Table~\ref{loss} displays, when the model is trained with all loss functions, the AUC is 59.88\% on DFDC. Nevertheless, a 4.47\% decrease of AUC could be realized by removing the EFD loss, demonstrating that expression-aware supervision helps the model capture high-level manipulation-relevant cues. Furthermore, the AUC of our model without the VFA loss $\mathcal{L}_{\mathrm{vfa}}$ drops by about 3.79\%, showing that vision-level contrastive regularization is beneficial for enhancing forgery representations. When the PPF loss is removed, the fairness gaps of our detector increase noticeably, and the AUC drops, demonstrating its importance in aligning subgroup distributions while preserving discriminability. Our model without the DSAD loss $\mathcal{L}_{\mathrm{dsad}}$ decreases by 1.26\% AUC, suggesting that authenticity class-aware language supervision can enhance vision-language fairness learning. When the group-class margin loss $\mathcal{L}_{\mathrm{mar}}$ is removed, the detection AUC of our detector decreases and the fairness gaps increase, indicating that explicitly separating real and fake prototypes across demographic groups is important for maintaining class discriminability. Similarly, removing the subgroup alignment loss $\mathcal{L}_{\mathrm{align}}$ results in worse fairness performance, especially on Celeb-DF and DFDC, which verifies its effectiveness in reducing demographic-induced feature shifts within the same authenticity class. The full loss achieves the best trade-off between fairness and generalization, validating the complementary effects of the proposed loss functions.

{\bfseries\setlength\parindent{0em} Influences of expression injector.}
In Table~\ref{loss}, the performance of FairForensics without EI decreases across unseen datasets than that of FairForensics with EI. Specifically, the AUC of our model with EI is increased by about 2.17\%, 2.09\%, and 0.96\% on FF++, Celeb-DF, and DFDC, respectively. Besides, EI also improves demographic fairness. For example, the $F_\text{FPR}$ and $F_\text{MEO}$ of our model with EI are reduced by 1.7\% and 1.06\% on Celeb-DF, respectively. We argue that the global EFP interaction adaptively injects expression-guided forgery cues into appearance features, pushing our detector to mine discriminative and generalizable forgery features.

\vspace*{-1em}
\subsection{Visualization Study}
{\bfseries\setlength\parindent{0em} Visualization of heatmaps.} 
To further understand the contribution of high-level EFP priors, we visualize the activation heatmaps of different fairness-enhanced detectors across various demographic subgroups. As shown in Fig.~\ref{heatmap}, PG-FDD and Ding et al. mainly focus on limited local regions such as background or low-level texture artifacts, and their attention patterns are relatively similar across different demographic faces. Without EFP, our model tends to focus on partial facial regions, which may still contain identity- or demographic-related cues. In contrast, as EFP are incorporated, our model attends to more diverse and semantically meaningful facial regions, such as eyes, mouth, facial contours, and expression-related areas. These areas are closely related to manipulation artifacts and are less dependent on specific demographic appearances. The heatmap visualizations indicate that expression-aware guidance helps our model capture more general and discriminative forgery traces, thereby improving both detection generalization and demographic fairness.

{\bfseries\setlength\parindent{0em} Visualization of prototype distributions.}  To intuitively analyze the effect of the PPF objective, we visualize the prototype distributions using t-SNE and the heatmap of intra-group and cross-group distances among prototypes. As shown in Fig.~\ref{prototype}, without the PPF objective, real and fake prototypes from different demographic subgroups are scattered irregularly, and the distances among subgroup prototypes show obvious variations. This indicates that the model may learn demographic-specific feature distributions, leading to biased decision boundaries across subgroups. By contrast, as the PPF objective is involved, prototypes belonging to the same authenticity class become more compact, while real and fake prototypes remain clearly separated. The heatmaps further show that intra-class subgroup distances are reduced and the cross-class subgroup distances are enlarged across demographic groups. This shows that our PPF objective can align subgroup distributions within the same class while maintaining discriminability across classes, thereby pushing the model to learn demographic-invariant forgery representations.
\vspace*{-1em}
\section{Conclusion}
In this paper, we present a novel framework for generalizable FDD that jointly models expression-aware visual forgery representations and demographic-aware semantic supervision. Specifically, we design an expression encoder and an expression-perceptual visual encoder to capture complementary forgery cues from both expression and global appearance, while mitigating identity-related biases via the plug-and-play identity-aware patch perturbation module. Furthermore, we introduce a demographic-guided language encoder to construct fine-grained vision-language alignment, enabling the model to disentangle forgery features from demographic attributes. To further enhance fairness, we propose a PPF objective that explicitly enforces inter-class separability and intra-class consistency across demographic subgroups. Extensive experiments on our balanced benchmark demonstrate that our method achieves state-of-the-art performance in terms of both generalization and fairness.

{\bfseries\setlength\parindent{0em} Limitations.}  While our method mitigates bias across predefined demographic groups, it does not explicitly address more complex, intersectional, or continuous attributes. As a result, residual bias may persist when multiple sensitive factors interact. More principled formulations could be investigated, such as causal modeling or invariant representation learning, to better disentangle forgery cues from confounding attributes.

\bibliographystyle{IEEEtran}

\bibliography{GazeCLIP}

%\begin{IEEEbiography}[{\includegraphics[width=1in,clip,keepaspectratio]{YaningZhang.png}}]{Yaning Zhang} received the double bachelor’s degree in Internet of Things Engineering and English and the M.S. degree in Computer Applied Technology from Qilu University of Technology (Shandong Academy of Sciences), Jinan, China, in 2020 and 2023, respectively, where she is currently pursuing the Ph.D. degree. Her research interests include computer vision, artificial intelligence, multimedia forensics, and face forgery detection.
%\end{IEEEbiography}

%\begin{IEEEbiography}[{\includegraphics[width=1in,clip,keepaspectratio]{ZanGao.png}}]{Zan Gao}  (Senior Member, IEEE) received his Ph.D degree from Beijing University of Posts and Telecommunications in 2011. He is currently a full Professor with the Shandong Artificial Intelligence Institute, Qilu University of Technology (Shandong Academy of Sciences). From Sep. 2009 to Sep. 2010, he worded in the School of Computer Science, Carnegie Mellon University, USA. From July 2016 to Jan 2017, he worked in the School of Computing of National University of Singapore. His research interests include artificial intelligence, multimedia analysis and retrieval, and machine learning. He has authored over 100 scientific papers in international conferences and journals including TPAMI, TIP, TNNLS, TMM, TCYBE, CVPR, ACM MM, WWW, SIGIR and AAAI.
%\end{IEEEbiography}

\vspace{-2em}
\end{document}